\pdfoutput=1

\documentclass[11pt]{article}

\usepackage[preprint]{acl}

\usepackage{times}
\usepackage{latexsym}
\usepackage{pdfpages}
\usepackage[T1]{fontenc}

\usepackage[utf8]{inputenc}

\usepackage{microtype}

\usepackage{inconsolata}

\usepackage{graphicx}

\usepackage{array}
\usepackage{colortbl}
\usepackage{xcolor}
\usepackage{tabularx}
\usepackage{booktabs}
\usepackage{multirow}
\usepackage{enumitem}
\usepackage{amsmath}
\usepackage{pifont}
\usepackage{xcolor}
\usepackage{multirow}
\usepackage{makecell}
\usepackage[most]{tcolorbox} 
\usepackage{siunitx}

\usepackage{setspace}
\usepackage{etoolbox}
\usepackage{algorithm}
\usepackage{algpseudocode} 
\usepackage{kotex}

\usepackage{caption}
\algrenewcommand\algorithmicindent{1em}

\newcommand{\cmark}{\textcolor{green!60!black}{\ding{51}}} 
\newcommand{\xmark}{\textcolor{red}{\ding{55}}} 

\title{Memorization or Reasoning? Exploring the Idiom Understanding of LLMs}

\author{
\textbf{Jisu Kim\textsuperscript{1\dag}},
\textbf{Youngwoo Shin\textsuperscript{1\dag}},
\textbf{Uiji Hwang\textsuperscript{1}},
\\
\textbf{Jihun Choi\textsuperscript{2}},
\textbf{Richeng Xuan\textsuperscript{3}},
\textbf{Taeuk Kim\textsuperscript{1*}}
\\
\textsuperscript{1}Hanyang University,
\textsuperscript{2}Sony AI,
\textsuperscript{3}Beijing Academy of Artificial Intelligence
\\
\texttt{\{jskim1945, youngwoo9, willpower, kimtaeuk\}@hanyang.ac.kr},
\\
\texttt{jihun.a.choi@sony.com}, \texttt{rcxuan@baai.ac.cn}
}

\begin{document}
\maketitle

\begingroup
\renewcommand{\thefootnote}{\fnsymbol{footnote}}
\footnotetext{\textsuperscript{\dag}Equal contribution. \textsuperscript{*}Corresponding author.}
\endgroup

\begin{abstract}
Idioms have long posed a challenge due to their unique linguistic properties, which set them apart from other common expressions.
While recent studies have leveraged large language models (LLMs) to handle idioms across various tasks, e.g., idiom-containing sentence generation and idiomatic machine translation, little is known about the underlying mechanisms of idiom processing in LLMs, particularly in multilingual settings.
To this end, we introduce MIDAS, a new large-scale dataset of idioms in six languages, each paired with its corresponding meaning. 
Leveraging this resource, we conduct a comprehensive evaluation of LLMs’ idiom processing ability, identifying key factors that influence their performance.
Our findings suggest that LLMs rely not only on memorization but also adopt a hybrid approach that integrates contextual cues and reasoning, especially when processing compositional idioms.
This implies that idiom understanding in LLMs emerges from an interplay between internal knowledge retrieval and reasoning-based inference.
\end{abstract}

\section{Introduction}

Idioms are a form of multi-word expression (MWE) in which a fixed combination of words functions as a single semantic unit.\footnote{Slang, proverbs, and sayings are representative examples of idioms, all of which are included in the scope of this study.}
Although their precise definition remains debated in linguistics, idioms are widely recognized as conventionalized expressions that often convey figurative meaning \cite{grant-2004}. 

\begin{figure}[t]
  \centering
  \includegraphics[width=0.86\linewidth]{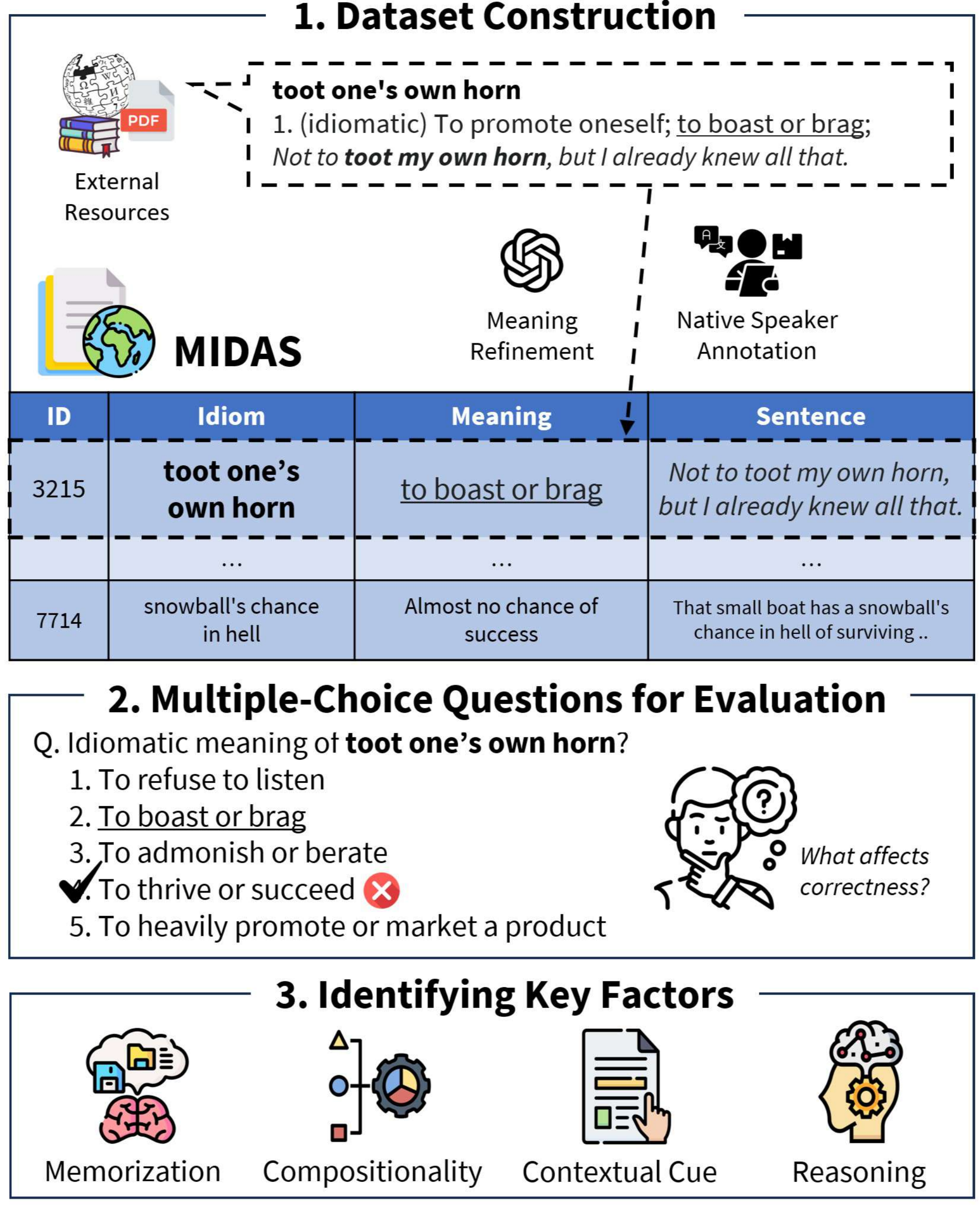}
  \caption{Evaluation framework for LLM idiom understanding: (1) building a large-scale multilingual idiom dataset (MIDAS), (2) designing multiple-choice questions for model assessment, and (3) evaluating performance across four key dimensions---memorization, compositionality, contextual cues, and reasoning.}
  \label{fig:main}
\end{figure}


Due to their unique characteristics that distinguish them from common expressions, idioms have long posed challenges in NLP \citep{sag-2002}. 
While many recent studies \citep{lee-2023, liu-2024, Li_2024, donthi-2025} have begun leveraging large language models (LLMs) \citep{Radford-2019, brown-2020} to tackle idiom-related tasks, there is growing concern that such approaches may be misguided: treating idioms as if they were ordinary expressions, without adequately accounting for their distinctive properties.
To promote responsible use of LLMs in idiom processing and establish best practices, it is crucial to assess both the extent of their idiom understanding and the factors that shape it.

However, evaluating idiom understanding in LLMs remains challenging due to limited resources \cite{liu-2024}.
The lack of resources is especially prominent in multilingual contexts, although accounting for idioms across languages is critically important due to their culture-specific nature. 

In addition, idioms are distinctive in that they lie at the intersection of fixed expressions---typically retrieved from memory \citep{Gibbs-1980}---and figurative expressions---interpreted compositionally through metaphorical reasoning \citep{GIBBS-1997}. 
Studies in linguistics suggest that humans process idioms in a hybrid manner, drawing on both characteristics \citep{hybrid}.
Yet, little is known about how LLMs process idioms along these dimensions, particularly whether they rely on memorization, reasoning, or both.

To address these issues, we introduce \textbf{MIDAS} (\textbf{M}ultilingual \textbf{I}diom \textbf{D}ataset \textbf{A}cross \textbf{S}ix Languages), a large-scale dataset of idioms in six languages, each paired with a carefully curated figurative meaning. 
As illustrated in Figure~\ref{fig:main}, we use MIDAS to conduct a comprehensive analysis of LLMs’ idiomatic competence and underlying mechanisms from multiple perspectives.

Specifically, we design a multiple-choice task in which the model selects the correct meaning of an idiom, serving as the primary means of evaluating idiom understanding. 
Building on this, we examine whether LLMs rely on memorization or compositional reasoning in processing idioms.

To assess reliance on memorization, we distinguish between memorized and unmemorized idioms using a continuation task, where the model is prompted to supply the final word of an idiom. 
We also evaluate the model’s use of compositionality---the degree to which meaning can be inferred from an idiom’s constituents---by grouping idioms based on the model's own estimation.

We take a step further by unveiling the influence of additional factors such as contextual cues and reasoning.
The role of context in idiom interpretation is examined by comparing performance with and without example sentences. 
To probe the role of reasoning, we conduct targeted analyses of recent models with strong reasoning capabilities.
Together, these experiments aim to uncover how LLMs interpret idioms across languages.

Experimental results reveal a significant performance gap between high-resource languages (e.g., English) and lower-resource ones (e.g., Korean).
We also find that LLMs adopt a hybrid approach, drawing on both memorization and compositionality. 
While sensitive to contextual cues, their use of reasoning remains inconsistent across languages.


\section{Related Work}

\subsection{Idiom Datasets}
Most idiom datasets are monolingual, focusing heavily on English \citep{saxena-2020, adewumi-2022, haagsma-2020, agrawal-2018} and Chinese \citep{zheng-2019, wu-2024, tang-2022}, while resources for low-resource languages remain scarce \citep{Wang-2024, shaikh-2024, igono2018translation, donthi-2025}, highlighting a significant disparity between languages.

While multilingual idiom datasets exist, they exhibit limitations. 
ID10M \cite{tedeschi-2022} omits idiom meanings, focusing solely on identification. 
LIdioms \cite{moussallem-2018} covers only European languages with a limited number of instances. 
MAPS \cite{liu-2024} similarly offers limited idiom instance coverage. IdiomKB \cite{Li_2024} relies entirely on GPT-3.5-generated meanings without human validation, raising concerns about reliability. 
To this end, we present MIDAS, an idiom dataset in six typologically diverse languages. 
Comparisons in Table \ref{tab:data} highlight the broader coverage and substantial size of MIDAS.

\begin{table}[t]
    \centering
    \scriptsize
    \setlength{\tabcolsep}{0.15em}
    \begin{tabular}{clc}
        \toprule
        \textbf{Datasets} & \textbf{\# Instances (Language)} & \textbf{Meaning}\\ 
        \midrule
        \multirow{2}{*}{\textbf{ID10M}} & 4,568 (EN), 1,301 (ZH) 1,229 (ES), 189 (NL), 188 (FR), & \multirow{2}{*}{\xmark} \\
        & 819 (DE), 452 (IT), 165 (JA),  648 (PL), 559 (PT) \\
        \midrule
        \textbf{LIdioms} & 291 (EN), 114 (PT), 175 (IT), 130 (DE), 105 (RU) & \cmark \\
        \midrule
        \multirow{2}{*}{\textbf{MAPS}} & 424 (EN), 364 (ZH), 364 (DE), 420 (RU), & \multirow{2}{*}{\cmark} \\
         & 370 (BN), 371 (ID) \\ 
        \midrule
        \textbf{IdiomKB} & 3,990 (EN), 8,643 (ZH), 270 (JA) & \textbf{\textcolor{orange}{$\Delta$}} \\ 
        \midrule
        \multirow{2}{*}{\makecell{\textbf{MIDAS} \\\textbf{(Ours)}}} & 9,766 (EN), 10,097 (DE), 11,851 (ZH), 11,316 (KO),  & \multirow{2}{*}{\cmark} \\
        & 8,051 (AR), 13,579 (TR) \\ 
        \bottomrule
    \end{tabular}
    \caption{Comparisons of multilingual idiom datasets, showing that MIDAS covers diverse languages, includes more instances, and provides corresponding meanings.
    $\Delta$: The meanings of the idioms are predicted by LLMs.}
    \label{tab:data}
\end{table}

\subsection{Idiom Processing of LLMs}
Limited work has examined key factors in LLMs’ idiom processing, such as memorization and compositionality, with existing studies offering conflicting views. \citet{liu-2024} suggest a weak correlation between memorization and processing. In contrast, \citet{walde-2024} argue that models rely heavily on memorization when processing MWEs, drawing on findings from noun compounds \citep{li-2022, coil-2023}. However, it remains unclear whether such patterns apply to idioms, given their distinct linguistic properties. \citet{khoshtab-2025} evaluate idioms and similes under zero-shot and CoT \citep{wei-2022} settings. They argue that MAPS-style\footnote{A binary classification task to identify the correct meaning of an idiom in context.} evaluation is insufficient and call for more challenging settings.
While we also evaluate LLMs’ idiom understanding, our work introduces a new approach that distinguishes from the MAPS format. 
We evaluate idiom understanding along with factors such as memorization and compositionality, analyzing their correlation with overall model performance.

\begin{figure*}[t!]
  \centering
  \includegraphics[width=0.9\textwidth]{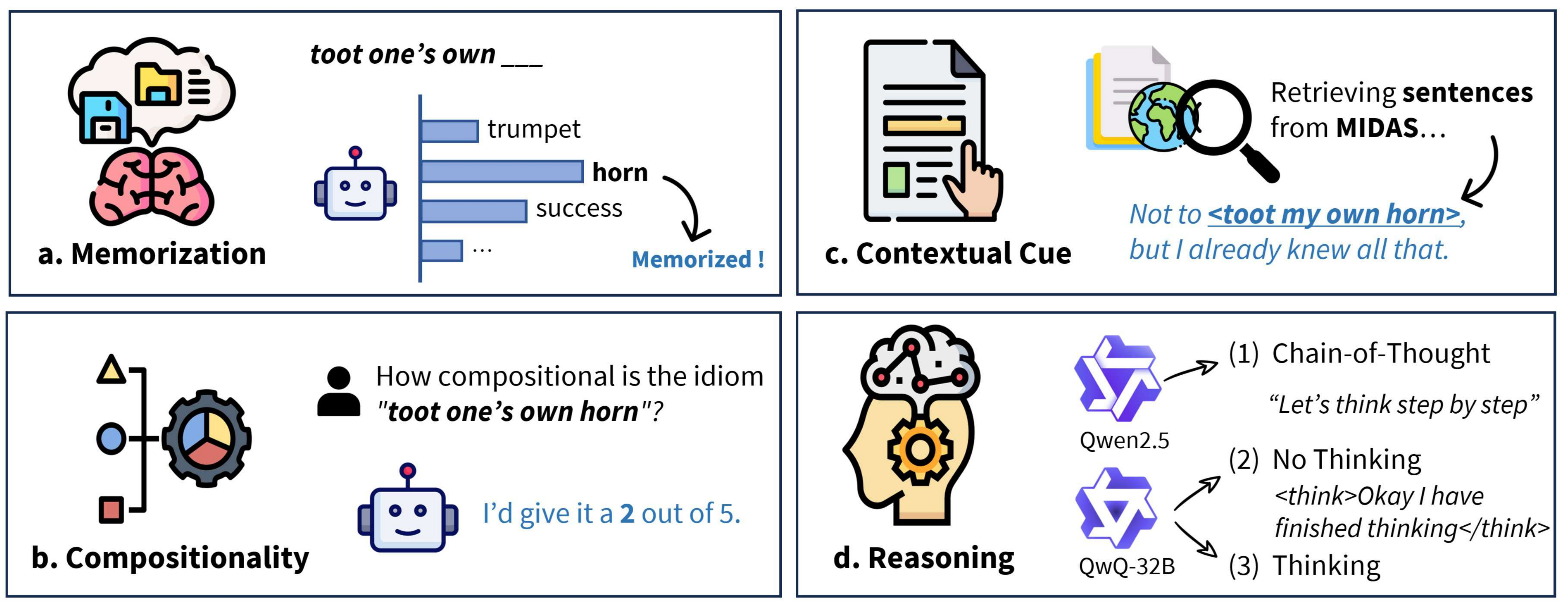}
  \caption{Visual illustration of four key factors---(a) memorization, (b) compositionality, (c) contextual cues, and (d) reasoning---that we control to examine their impact on idiom understanding in LLMs.}
  \label{fig:Factors}
\end{figure*}

\section{MIDAS: A Multilingual Idiom Dataset Across Six Languages}
\label{sec:midas}

To overcome the limitations of existing datasets and establish a robust foundation for multilingual idiom research, we present \textbf{MIDAS} (\textbf{M}ultilingual \textbf{I}diom \textbf{D}ataset \textbf{A}cross \textbf{S}ix Languages),\footnote{Available at: \url{https://github.com/HYU-NLP/MIDAS}} a comprehensive dataset spanning six typologically and culturally diverse languages: English (EN), German (DE), Chinese (ZH), Korean (KO), Arabic (AR), and Turkish (TR).
MIDAS is a large-scale dataset containing 64,660 idiomatic expressions, each paired with a figurative meaning. 
Where available, example sentences are also included, refined through a process to minimize noise and ensure high quality.
Figure \ref{fig:main} visualizes an example from the English subset of MIDAS. 
Statistics and representative examples of MIDAS can be found in Appendix \ref{appendix_A.1} and \ref{appendix_A.2}.

\paragraph{Dataset construction}
Given the limited availability of well-structured idiom datasets, particularly in multilingual contexts, the first step is to collect data from diverse sources via web scraping and extraction from PDF-based e-books. In this process, we prioritize sources that: (1) cover a sufficiently large number of instances, (2) provide detailed meanings for each idiom, including example sentences where available, and (3) are produced by recognized and reliable entities, preferably authoritative institutions for the given language. 
Consequently, we take \textit{Wiktionary}, \textit{Duden–Redewendungen}, the \textit{Xinhua} dictionary, the \textit{Korean Standard Dictionary}, \textit{A Dictionary of Arabic Idioms and Expressions}, and the \textit{Turkish Idioms and Proverbs} dataset as sources for English, German, Chinese, Korean, Arabic, and Turkish, respectively. 

Next, we apply language-specific preprocessing to remove noise and address issues from the data sources or collection processes.\footnote{For example, in the English dataset, some meanings appear as ``Alternative form of \textit{idiom}'', which we treat as variations of the referenced \textit{idiom} rather than as separate instances. In the Turkish dataset, we separate all example sentences from their corresponding meanings, since they were not stored in distinct columns in the original data.}
We then refine the meanings to better align with our objectives, as follows.
First, we make use of LLMs to automatically refine meanings by removing extraneous details and noise---such as lengthy explanations of how an idiom’s surface form relates to its meaning or descriptions of its origin---while preserving the core semantics.
We achieve this by prompting LLMs with only the meaning of each idiom, excluding its surface form, to minimize bias from prior familiarity or knowledge. 
Specifically, we utilize GPT-4o for all languages, except for English, where GPT-4o mini is sufficient. 
The automated refinement is then followed by native-speaker annotation to ensure authenticity and accuracy. 
This step is critical for maintaining high-quality meanings, which serve as gold labels throughout experiments.

Furthermore, unlike many existing datasets that provide only a single form-meaning pair per idiom, we preserve all surface variations and semantic nuances.
Form variants are stored as lists, while each distinct sense is assigned its own row. 
Further details on data sources, extraction, and preprocessing are provided in Appendix \ref{appendix_A.3}, while legal and ethical considerations are discussed in Appendix \ref{appendix_A.4}.

\section{Methodology} \label{sec:methdology}
Using MIDAS, we design a series of evaluations to investigate the idiomatic competence of LLMs. The central task is to query LLMs in a multiple-choice question format, where the model is requested to select the correct meaning of a given idiom. 
We further present auxiliary analyses to identify factors influencing idiom comprehension, examining whether performance is driven by memorization or compositionality, along with the impact of contextual understanding and reasoning ability.

\subsection{Multiple-Choice Questions (MCQs)}
Previous studies, e.g., DICE and MAPS \citep{mi-2024, liu-2024, khoshtab-2025}, chiefly rely on binary classification, where models are tasked with identifying the correct meaning of an idiom in context from two candidates.
However, this approach lacks scalability, as it relies on manually crafted problem sets.
Furthermore, it poses a barrier to isolating and evaluating the specific contribution of context to idiom understanding.
To this end, we adopt multiple-choice questions (MCQs) \citep{hendrycks-2021, zhang-2024, wang-2025} as our evaluation framework, based on the observation that idioms possess fixed, conventionalized meanings, which allow for an evaluation style akin to factual knowledge assessment.

\paragraph{MCQ creation} 
We employ five-option MCQs, each comprising one correct meaning and four distractors: two derived from the idiom’s surface form and two from its figurative meaning.
The goal is to test whether LLMs can identify the true meaning of an idiom or are overly influenced by surface-level or semantically related distractors.
All options are drawn from MIDAS, whose wide coverage enables the selection of plausible and challenging ones.

The algorithm selects answer options based on their similarity to the surface form and meaning of the target idiom,\footnote{To prevent multiple correct or ambiguous choices, we exclude candidates that rank in the top 1\% in similarity for both surface-form-based and meaning-based options.} based on cosine similarity between sentence embeddings.\footnote{We employ \texttt{\scriptsize intfloat/multilingual-e5-large-instruct} as our multilingual sentence embedding model, which is proven to excel in semantic similarity tasks \citep{enevoldsen-2025}. The model is available at \url{https://huggingface.co/intfloat/multilingual-e5-large-instruct}.}
The pseudocode for MCQ construction is provided in Algorithm \ref{alg:mcq_pseudocode}.

\subsection{Continuation as Idiom Memorization} \label{subsection 4.2}
Unlike other expressions, idioms are lexically and syntactically fixed, i.e., changes in wording or structure often obscure their meaning or render them unnatural.
Motivated by this, prior studies \citep{haviv-2023, liu-2024} suggest that memorized and unmemorized idioms can be distinguished by testing whether a model can predict an idiom’s final word given its preceding context.
For example, if a model can predict \textit{``lining''} given \textit{``every cloud has a silver''}, it suggests that the idiom is memorized.
We adopt a similar continuation approach to distinguish between memorized and unmemorized idioms.

\paragraph{Filtering continuation candidates}
Not all idioms lend themselves to continuation-based evaluation. 
Some contain words that are trivially predictable or appear in contexts allowing multiple valid completions.
Inspired by \citet{haviv-2023}, we apply the filtering rules detailed in Appendix~\ref{appendix_continuation_filter} to exclude unsuitable idioms.
We use only filtered idioms in the following, with their statistics reported in Appendix~\ref{appendix_continuation_candidate}.

\paragraph{Grouping by memorization}
We classify the filtered idioms into memorized and unmemorized groups.
We input an idiom---excluding the last target word---in the user prompt, with the temperature set to 0.\footnote{We observe that closed-source models tend to generate unnecessary outputs when given only idiom expressions. To address this, we append the instruction \textit{``You are a next-word prediction engine. Only return the single next word to complete the given expression''} in the system prompt.}  
Since closed-source models do not officially support greedy decoding and may produce non-deterministic outputs even at temperature 0, we run each idiom five times and check whether at least one output starts with the target word’s first token. For open-source models, we verify whether the first token of the target word appears among the top five candidates by log probability.
Finally, an idiom is labeled as memorized if it passes the test; otherwise, it is considered unmemorized.
In \S\ref{subsec:memorization}, we utilize the classified idioms to analyze the impact of memorization on idiom processing.

\subsection{Measuring Compositionality Scores} \label{subsection 4.3}
While idioms are often characterized by their non-compositionality, the extent to which their meanings can be inferred from their constituents varies significantly. 
For instance, idioms such as \textit{``kick the bucket''} offer little semantic transparency, while others such as \textit{``every cloud has a silver lining''} exhibit partially inferable meanings through metaphorical reasoning. 
To probe this distinction, we define compositionality scores for each idiom, reflecting the degree to which its meaning is predictable from its constituents, in order to evaluate whether models engage in metaphorical reasoning when processing idioms.

Concretely, we utilize a prompt-based approach to elicit compositionality judgments directly from LLMs. 
Models are presented with an idiom, its meaning, and a definition of compositionality, and asked to rate how inferable the meaning is from its components on a 1–5 scale.
For example, given the idiom ``\textit{paint a rosy picture}'' and its meaning (``\textit{to describe a situation optimistically}''), the model is requested to assign a score based on the contribution of its components---e.g., \textit{paint}, \textit{rosy}, and \textit{picture}. 
A score of 5 indicates high compositionality, while a score of 1 denotes complete opacity.
This method offers a scalable and interpretable way to approximate compositionality without relying on idiom embeddings or other complex techniques.\footnote{It also allows us to examine whether the model’s own perception of compositionality aligns with its actual idiom comprehension, as measured by performance on MCQ tasks.} 
Refer to Appendix \ref{appendix_Comp} for additional details and statistics on compositionality scores.

\subsection{Models}
We employ open- and closed-source models trained to support the six languages covered in MIDAS.
We treat GPT-4o \cite{openai-2024}, DeepSeek-V3 \cite{deepseekai-2025} as closed-source,\footnote{Although DeepSeek is open-source, we use its API due to resource constraints, limiting control to closed-source levels.} Aya-Expanse-32B \cite{dang-2024} and Qwen2.5-32B \cite{qwen-2025} as open-source. 
We also include QwQ-32B \cite{qwq32b}, a reasoning-enhanced Qwen2.5-32B, for rigorous comparison.


\subsection{MCQ Evaluation}
\paragraph{Experimental setups} 
We use localized prompts that ask for idiomatic meanings in each language, phrased naturally and verified by native speakers. 
MCQs are tested in a zero-shot setting.
To ensure robustness in evaluation, we enforce a strict protocol: each MCQ appears three times with shuffled choices, placing the correct answer in a new position each time. 
A model is deemed correct only if it selects the right answer in all three trials. 


\begin{table}[t]
\centering
\scriptsize
\setlength{\tabcolsep}{5pt}
\begin{tabular}{lcccccc}
\toprule
\textbf{Model} & \textbf{EN} & \textbf{DE} & \textbf{ZH} & \textbf{KO} & \textbf{AR} & \textbf{TR} \\
\midrule
Aya-Expanse-32B & 81.71 & 71.77 & 75.45 & 49.89 & 65.62 & 48.94 \\
Qwen2.5-32B & 83.71 & 73.94 & 93.35 & 51.39 & 71.25 & 40.31 \\
DeepSeek-V3 & 90.34 & 83.94 & \textbf{95.65} & 55.64 & \textbf{75.53} & 62.52 \\
GPT-4o & \textbf{91.13} & \textbf{88.08} & 91.44 & \textbf{72.72} & 72.85 & \textbf{71.82} \\
\bottomrule
\end{tabular}
\caption{Accuracy (\%) of LLMs on MCQs constructed from MIDAS. Best scores per language are in \textbf{bold}.}
\label{tab:mcq_results}
\end{table}

\begin{table}[t]
  \centering
  \scriptsize
  \setlength{\tabcolsep}{5pt}
  \begin{tabular}{l c c c c c c}
    \toprule
    \textbf{Model} & \textbf{EN} & \textbf{DE} & \textbf{ZH} & \textbf{KO} & \textbf{AR} & \textbf{TR} \\
    \midrule
    Aya-Expanse-32B & \textbf{80.36} & 56.43 & \textbf{92.95} & \textbf{36.59} & \textbf{30.54} & 32.66 \\
    Qwen2.5-32B     & 73.72 & 45.27 & 77.97 & 31.61 & 29.87 & 22.28 \\
    DeepSeek-V3     & 70.83 & \textbf{59.28} & 89.51 & 31.06 & 29.45 & \textbf{45.82} \\
    GPT-4o          & 67.18 & 49.53 & 70.26 & 26.13 & 27.25 & 35.08 \\
    \bottomrule
  \end{tabular}
  \caption{Memorization rate (\%) across languages and models. Best scores per language are in \textbf{bold}.}
  \label{tab:mem}
\end{table}
\section{Experiments}
In this section, we delve into a series of experiments based on the MIDAS dataset described in \S\ref{sec:midas} and the analysis techniques outlined in \S\ref{sec:methdology}.
In particular, we probe how LLMs process idioms with respect to memorization, compositionality, context, and reasoning, as illustrated in Figure~\ref{fig:Factors}.

\paragraph{MCQ as a diagnostic framework}
MCQs serve as our primary tool for evaluating idiom understanding, applied across various cases tailored to test specific factors. 
Overall performance on the full MIDAS dataset is reported in Table~\ref{tab:mcq_results}. 
All models show a clear performance divide: they excel on English, German, and Chinese but fall behind on Korean, Arabic, and Turkish.
While the overall ranking is GPT-4o > DeepSeek-V3 > Qwen2.5 > Aya-Expanse, there are exceptions, notably in Chinese, where Qwen2.5 outperforms GPT-4o. 
These results suggest that, for idiom processing, choosing a language-specific model can be better than relying on a single model across all languages.

\begin{table}[t]
\centering
\scriptsize
\begin{tabular}{lcccl}
\toprule
\textbf{Model} & \textbf{Lang.} & \textbf{Acc. (\cmark)} & \textbf{Acc. (\xmark)} & \textbf{$\Delta$ Acc.} \\
\midrule
\multirow{6}{*}{\texttt{Aya-expanse-32B}}
                         & EN & 86.93 & 73.64 & 13.29$^{***}$ \\
                         & DE & 76.47 & 67.32 & 9.15$^{**}$ \\
                         & ZH & 76.25 & 61.66 & 14.59$^{***}$ \\
                         & KO & 60.13 & 46.62 & 13.51$^{***}$ \\
                         & AR & 72.55 & 64.71 & 7.84$^{*}$ \\
                         & TR & 60.57 & 51.20 & 9.37$^{**}$ \\
\midrule
\multirow{6}{*}{\texttt{Qwen2.5-32B}} & EN & 93.40 & 77.57 & 15.83$^{***}$ \\
                         & DE & 82.99 & 70.09 & 12.90$^{***}$ \\
                         & ZH & 92.08 & 87.83 & 4.25$^{*}$ \\
                         & KO & 62.17 & 51.17 & 11.00$^{***}$ \\
                         & AR & 78.89 & 66.72 & 12.17$^{***}$ \\
                         & TR & 54.40 & 44.43 & 9.97$^{***}$ \\
\midrule
\multirow{6}{*}{\texttt{DeepSeek-V3}} & EN & 95.74 & 91.04 & 4.70$^{***}$ \\
                         & DE & 86.49 & 77.39 & 9.10$^{***}$ \\
                         & ZH & 96.92 & 93.10 & 3.82$^{**}$ \\
                         & KO & 67.11 & 54.33 & 12.78$^{***}$ \\
                         & AR & 81.94 & 69.02 & 12.92$^{***}$ \\
                         & TR & 73.86 & 60.65 & 13.21$^{***}$ \\
\midrule
\multirow{6}{*}{\texttt{GPT-4o}} & EN & 96.44 & 90.42 & 6.02$^{***}$ \\
                         & DE & 92.20 & 85.64 & 6.56$^{***}$ \\
                         & ZH & 93.43 & 80.03 & 13.40$^{***}$ \\
                         & KO & 83.31 & 72.91 & 10.40$^{***}$ \\
                         & AR & 80.98 & 73.87 & 7.11$^{**}$ \\
                         & TR & 85.91 & 72.37 & 13.54$^{***}$ \\
\bottomrule
\end{tabular}
\caption{
MCQ accuracy for memorized (\cmark) vs.\ unmemorized (\xmark) idioms.
$\Delta$ Acc. is marked with $^*$ ($p{<}.05$), $^{**}$ ($p{<}.01$), and $^{***}$ ($p{<}.001$) based on Fisher's exact test.
The results reveal that memorization has a statistically significant impact on idiom processing.
}
\label{tab:memorization_effect}
\end{table}

\subsection{Memorization}
\label{subsec:memorization}


\paragraph{Memorization rate}
We first report the extent to which LLMs memorize idioms, as estimated by the method presented in \S\ref{subsection 4.2}.
Table~\ref{tab:mem} indicates that models generally memorize idioms more in English, German, and Chinese than in Korean, Arabic, and Turkish.
For most cases, Aya-Expanse demonstrates the highest memorization rate, while GPT-4o shows the lowest---except in Turkish.
This implies that model size may not positively correlate with idiom memorization performance.


\paragraph{Memorization affects idiom understanding}
We conduct experiments using the two groups of idioms prepared in \S\ref{subsection 4.2}. 
As shown in Table~\ref{tab:mem}, the memorized and unmemorized groups are imbalanced; to address this, we sample balanced subsets per group (see Appendix~\ref{appendix_memorization_sample} for sampling details).

Table \ref{tab:memorization_effect} shows that LLMs achieve notably higher accuracy on memorized idioms. 
However, the magnitude of this effect varies across models and languages. 
Qwen2.5-32B exhibits the largest gaps, often exceeding 10\% points, peaking at +15.8\% in English. 
GPT-4o and DeepSeek-v3 also benefit from memorization, though to a lesser degree.

By language, Korean and Turkish consistently show large gains from memorization (e.g., +13.5\% in Turkish for GPT-4o, +12.8\% in Korean for DeepSeek-v3), suggesting that performance in these languages is more strongly driven by the ability to recall idioms.
In contrast, memorization appears to have a smaller effect in Chinese, particularly for Qwen2.5-32B and DeepSeek-v3, where accuracy is high even on unmemorized idioms. 
This suggests that models may rely on other capabilities, such as compositional interpretation, contextual integration, or broader linguistic generalization, to compensate for the lack of direct recall.

In summary, memorization serves as a useful shortcut for idiom understanding, although its importance varies by condition. 
When memorization is unavailable, LLMs may instead draw on their general abilities, such as contextual reasoning and semantic composition, explored in the following.

\begin{table}[t]
\centering
\scriptsize
\begin{tabular}{lcccc}
\toprule
\textbf{Model} & \textbf{Lang.} & \textbf{CS (\cmark)} & \textbf{CS (\xmark)} & \textbf{$\Delta$ CS} \\
\midrule
\multirow{6}{*}{\texttt{Aya-expanse-32B}} & EN & 3.00 & 2.65 & 0.35 (13.21\%) \\
                         & DE & 3.13 & 3.00 & 0.13 (4.33\%) \\
                         & ZH & 3.14 & 3.00 & 0.14 (4.67\%) \\
                         & KO & 3.00 & 2.92 & 0.08 (2.74\%) \\
                         & AR & 3.56 & 3.25 & 0.31 (9.54\%) \\
                         & TR & 3.12 & 3.07 & 0.05 (1.63\%) \\
\midrule
\multirow{6}{*}{\texttt{Qwen2.5-32B}} & EN & 2.40 & 1.68 & 0.72 (42.86\%) \\
                         & DE & 1.92 & 1.49 & 0.43 (28.86\%) \\
                         & ZH & 2.25 & 1.59 & 0.66 (41.51\%) \\
                         & KO & 1.52 & 1.22 & 0.30 (24.59\%) \\
                         & AR & 2.52 & 1.75 & 0.77 (44.00\%) \\
                         & TR & 1.78 & 1.53 & 0.25 (16.34\%) \\
\midrule
\multirow{6}{*}{\texttt{DeepSeek-V3}} & EN & 2.87 & 2.29 & 0.58 (25.33\%) \\
                         & DE & 2.62 & 2.21 & 0.41 (18.55\%) \\
                         & ZH & 2.87 & 2.50 & 0.37 (14.80\%) \\
                         & KO & 2.33 & 2.05 & 0.28 (13.66\%) \\
                         & AR & 2.72 & 2.38 & 0.34 (14.29\%) \\
                         & TR & 2.48 & 2.19 & 0.29 (13.24\%) \\
\midrule
\multirow{6}{*}{\texttt{GPT-4o}} & EN & 2.69 & 2.09 & 0.60 (28.71\%) \\
                         & DE & 2.45 & 1.99 & 0.46 (23.12\%) \\
                         & ZH & 2.18 & 1.86 & 0.32 (17.20\%) \\
                         & KO & 1.95 & 1.63 & 0.32 (19.63\%) \\
                         & AR & 3.17 & 2.78 & 0.39 (14.03\%) \\
                         & TR & 2.20 & 1.96 & 0.24 (12.24\%) \\
\bottomrule
\end{tabular}
\caption{
Average compositionality scores (CS) on a 1–5 scale for idioms answered correctly (\cmark) vs.\ incorrectly (\xmark). $\Delta$ CS: the absolute difference with relative increase (\%) in parentheses, indicating how compositionality perception correlates with idiom comprehension success.
}
\label{tab:compositionality_effect}
\end{table}

\subsection{Compositionality}
In this part, we employ the compositionality scores defined in \S\ref{subsection 4.3} to estimate the influence of compositional reasoning in LLMs. 
The core assumption is that a positive correlation between performance and compositionality scores indicates that LLMs are leveraging compositional reasoning.

\paragraph{LLMs exploit compositional reasoning}
Idioms are generally considered non-compositional expressions, as reflected in the compositionality scores assigned by LLMs, which tend to cluster toward the lower end of the 1–5 scale (typically between 1.5 and 3.0).
Despite this trend, idioms answered correctly consistently receive higher compositionality scores than those answered incorrectly. 
These differences are statistically significant, as confirmed by the Mann–Whitney U test \citep{Mann-Whitney} across all model–language combinations.

To examine this more closely, we present Table~\ref{tab:compositionality_effect}, which compares compositionality scores (CS) between the correct and incorrect response groups.
Among all models, Qwen2.5-32B displays the most distinct separation between the two groups. 
In Arabic and English, correctly answered idioms obtain scores 44\% (+0.769) and 43\% (+0.715) higher than incorrect ones.
In contrast, Aya-Expanse displays a much weaker pattern; in Turkish, the corresponding increase is only 1.6\% (+0.049).

Among the six languages, English and Arabic consistently show the largest compositionality effects. 
In English, for instance, the difference in average scores between correctly and incorrectly answered idioms exceeds 0.6 for both GPT-4o and DeepSeek-v3, and reaches 0.715 for Qwen2.5-32B. Arabic similarly shows strong effects, with Qwen2.5-32B exhibiting a 44\% increase in compositionality scores for correctly answered idioms. 
These results imply that in certain languages, compositionality acts as a more salient signal for successful idiom interpretation.

Taken together, these findings suggest that models are more likely to correctly interpret idioms they internally perceive as more compositional---those whose meanings are more readily inferable from their constituents. 
This consistent pattern across models and languages highlights compositional signals as a useful heuristic for idiom interpretation.


\begin{figure}[t]
    \centering
    \includegraphics[width=0.85\linewidth]{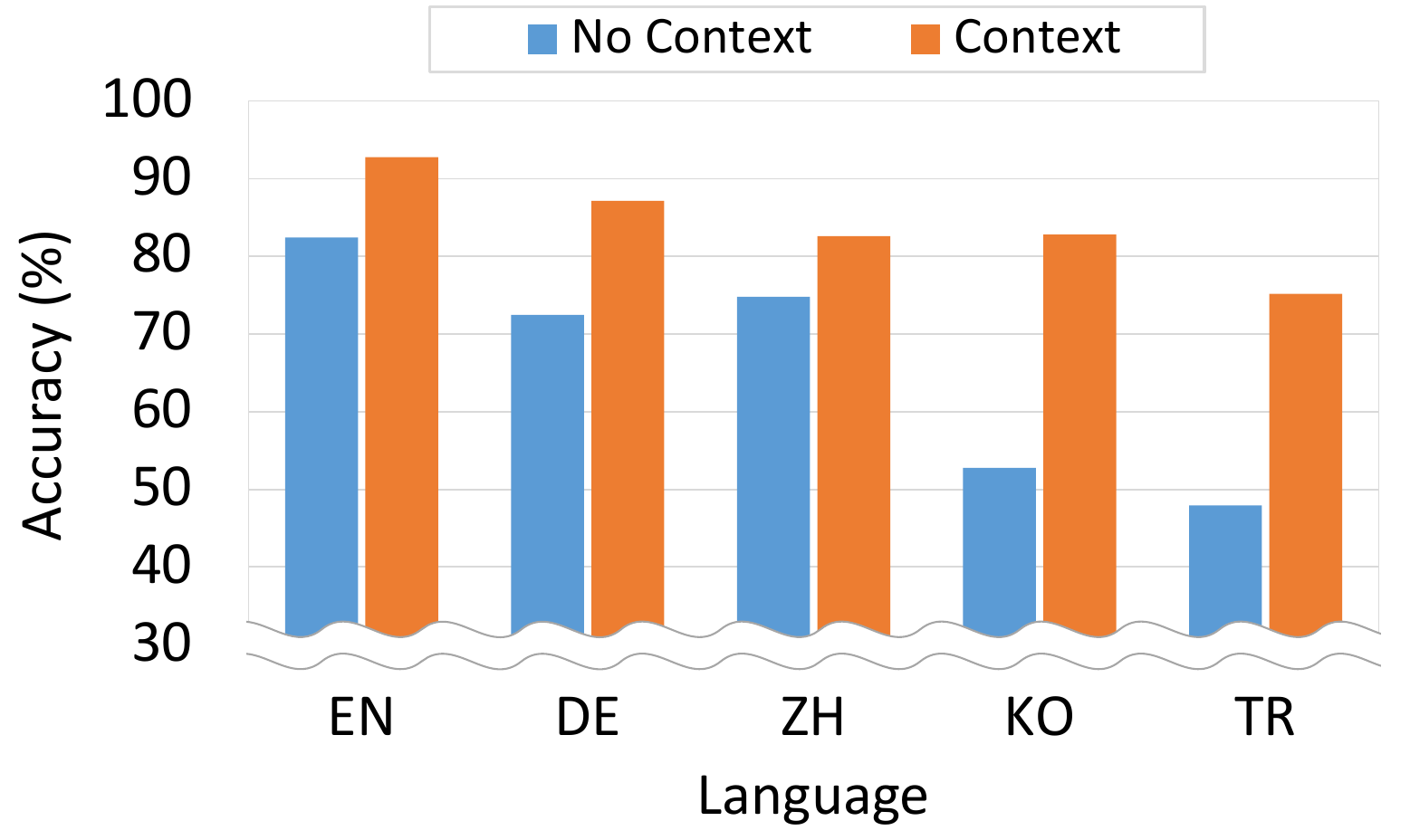}
    \caption{Accuracy comparison with and without context across languages for Aya-Expanse-32B.}
    \label{fig:context_effect}
\end{figure}

\subsection{Context} 
\label{subsec:context}

Having observed that LLMs tend to perform better on idioms that are either memorized or more compositional, a natural question arises: how do models leverage further contextual cues to infer idiomatic meanings? 
To explore this, we investigate whether providing usage examples can enhance model performance in the MCQ task.

We compare MCQ accuracy with and without example sentences drawn from MIDAS, restricting our analysis to idioms with annotated examples. 
Arabic is excluded, as it lacks example sentences (see Table~\ref{tab:stat} for details).
To match dataset sizes across languages, we downsample all datasets to approximately 3,700 instances per language.

Figure \ref{fig:context_effect} illustrates the effect of context---i.e., example sentences---for Aya-Expanse-32B, revealing that providing context substantially improves performance. 
This trend is consistent across all models we evaluated, demonstrating that LLMs are indeed capable of using contextual information to interpret idiomatic expressions more accurately.
Notably, Korean (KO) and Turkish (TR) exhibit the largest gains in accuracy, highlighting the potential of context to compensate for limited idiom exposure in low-resource language settings.
Detailed configurations for the experiment are provided in Appendix~\ref{appendix_ContextPrompt}.
Full results for all models are presented in Table~\ref{tab:context_results_all_models} in Appendix~\ref{appendix_context}.


\begin{figure}[t!]
  \centering
  \includegraphics[width=0.9\linewidth]
    {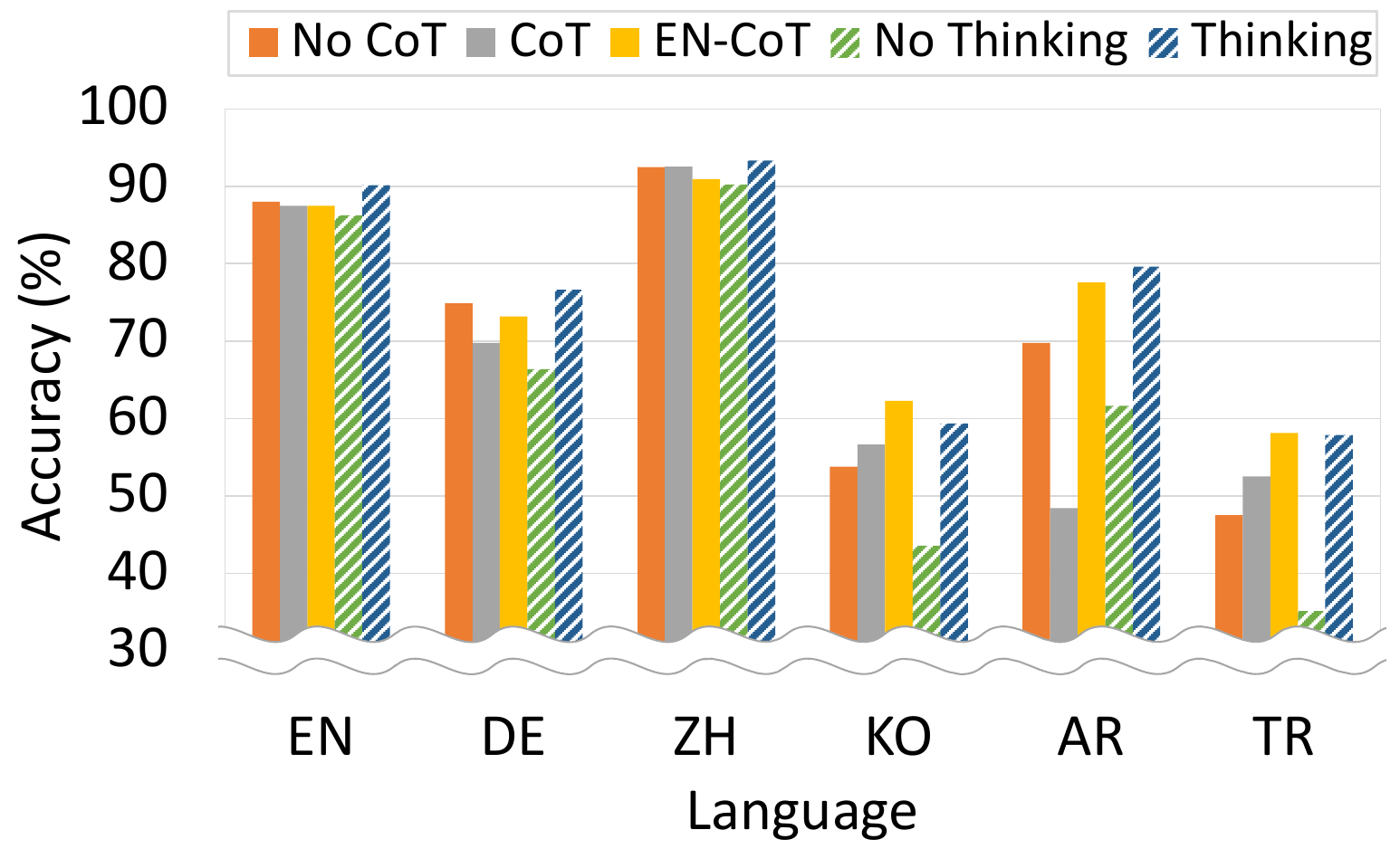}
  \caption{
  Comparison of various reasoning methods applied to two Qwen variants, including QwQ, which is specialized for reasoning.
  \textbf{No-CoT} shows Qwen’s performance without chain-of-thought (CoT) prompting, while \textbf{CoT} and \textbf{EN-CoT} represent results with CoT applied in the native language and in English, respectively. 
  \textbf{NoThinking} and \textbf{Thinking} denote QwQ’s performance without and with its built-in reasoning capabilities.
  }
  \label{fig:reasoning}
\end{figure}

\subsection{Reasoning}\label{main:reasoning}

Since LLMs leverage figurative thinking and contextual cues, it is also plausible that reasoning could enhance the performance of idiom processing.
To probe this question, we compare Qwen2.5 with its reasoning-augmented variant, QwQ, examining whether reasoning improves idiom understanding.
We compare several conditions: (1) Qwen2.5 with and without chain-of-thought (CoT) prompting, denoted as \textbf{CoT} and \textbf{No-CoT}, respectively.\footnote{We add the translated phrase ``Think step by step before you answer'' to activate zero-shot CoT \citep{kojima-2022}.} 
(2) Qwen2.5 with the English zero-shot CoT prompt (\textbf{EN-CoT}). 
(3) QwQ with and without reasoning, denoted as \textbf{Thinking} and \textbf{NoThinking}.\footnote{We rely on the recent NoThinking \citep{ma-2025} method to control the reasoning behavior of QwQ.}
This setup facilitates direct comparison of distinct reasoning strategies across LLMs, as well as the impact of language in the reasoning process.

The overall results are shown in Figure~\ref{fig:reasoning}.
We observe that the performance gain of CoT is not consistent across languages, which is similar to the findings of \citet{khoshtab-2025}. 
Specifically, performance drops are observed for English, German, Chinese, and Arabic, while Korean and Turkish show improvements. 
EN-CoT outperforms CoT in all languages except Chinese, but still falls short of No-CoT in English, German, and Chinese. 
For Arabic, where native CoT exhibits a noticeable performance decline, we find that reasoning paths for incorrect instances are often unstable, exhibiting issues such as code-mixing and refusal to answer (see Appendix~\ref{appendix_arabic} for details). 
This hints that the sharp drop in performance may be due to the model’s limited proficiency in Arabic.

The outcomes offer practical guidance for applying CoT to idiom processing. 
In particular, CoT appears most beneficial for lower-performing languages when reasoning is conducted in English, contrary to the common expectation that reasoning in a language’s native form would yield greater gains.
However, it should be noted that CoT may hinder performance in languages unfamiliar to the model, such as Arabic, as well as in languages that already perform strongly, such as English.

On the other hand, Thinking steadily outperforms all other settings across languages, while notably, NoThinking performs worse not only than Thinking but also than No-CoT. 
This stands in contrast to the findings of \citet{ma-2025}, which report NoThinking outperforming Qwen2.5-Instruct on seven math- and coding-related benchmarks. 
This discrepancy implies that although reasoning models gain enhanced mathematical and coding capabilities, these improvements may entail trade-offs resembling catastrophic forgetting of knowledge in continual learning.
We provide additional experimental results in Appendix~\ref{appendix_additional_reasoning}, which reveal similar language-specific inconsistencies and show that reasoning-augmented models lag behind their original counterparts.

\subsection{Factor Correlation Analysis}\label{main:factors}
The previous experiments isolate each factor’s effect on idiom comprehension but leave their interactions unclear. 
For instance, additional context may be more helpful for idioms with low compositionality or those not memorized.
To explore this, we analyze pairwise correlations of the four key factors, omitting the context–reasoning case, i.e., ${4\choose 2}$-1=5 combinations.\footnote{5 cases: memorization-compositionality, memorization-context, memorization-reasoning, compositionality-context, compositionality-reasoning.}
Specifically, we compute accuracy gains for two reasoning-related pairs---defined as the difference between CoT and No-CoT---and for two context-related pairs---defined as the difference between with- and without-context performance.
For memorization–compositionality, we examine the correlation between compositionality scores and memorization rates.

\begin{figure}[t]
  \centering
  \includegraphics[width=0.9\linewidth]{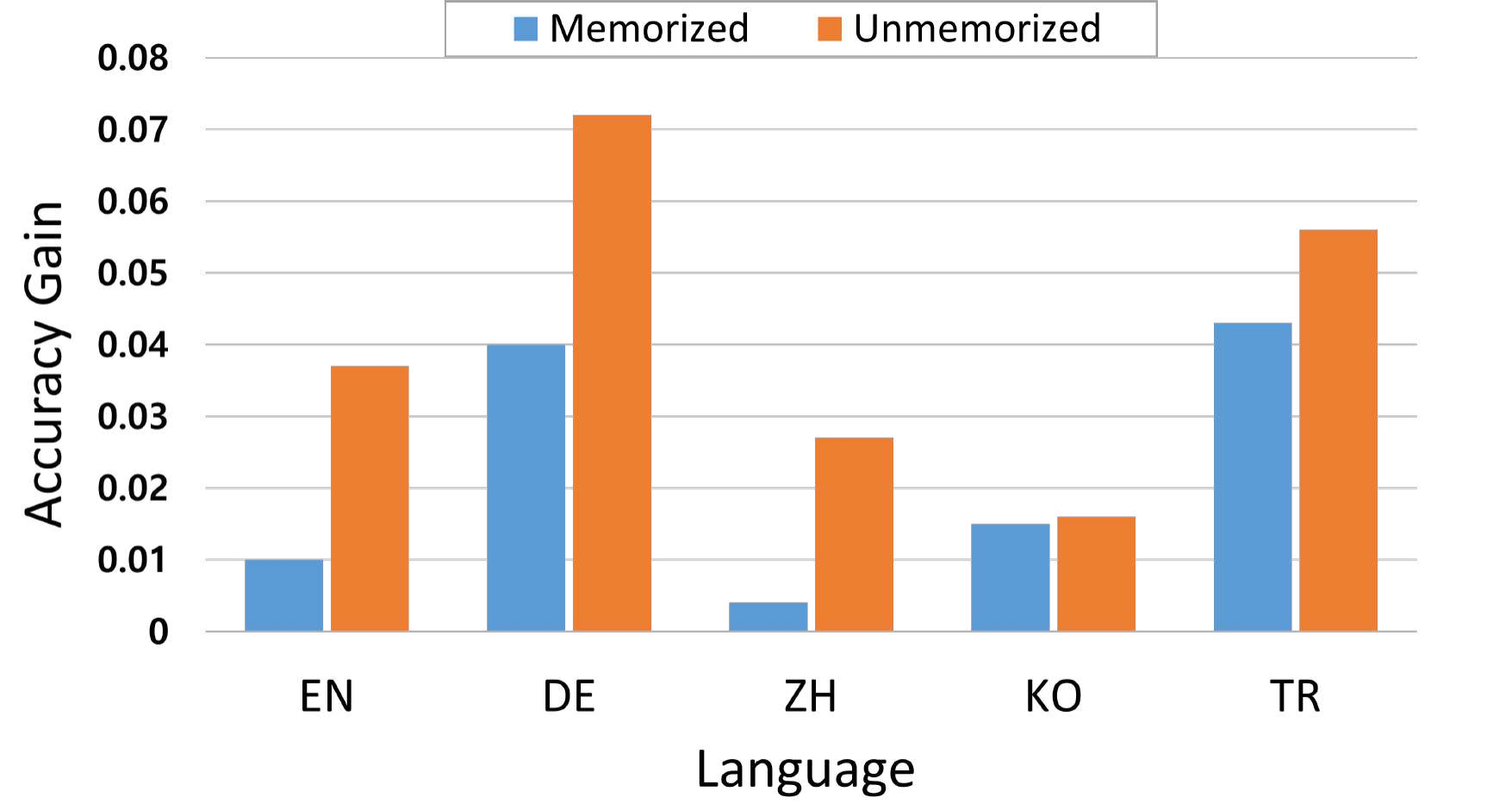}
  \caption{Accuracy gains from provided context (i.e., the difference in performance with vs. without context) are more pronounced when LLMs process unmemorized idioms. These results are based on Qwen2.5-32B-Instruct.}
  \label{fig:mem-cont}
\end{figure}

As a result, we observe patterns indicating that higher compositionality may be associated with easier memorization. 
Moreover, memorization and context are correlated: models benefit more from context when handling idioms they have not memorized (see Figure~\ref{fig:mem-cont}), implying that contextual cues help compensate for gaps in parametric knowledge.
In contrast, no notable correlations emerge for memorization–reasoning or compositionality–reasoning, partly due to language-specific inconsistencies, and compositionality–context likewise shows no clear relationship. 
Additional details and figures are provided in Appendix~\ref{appendix_factor}.

\section{Revisiting Prior Work with Findings}
Recent studies have leveraged LLMs for idiom-related tasks such as example sentence generation \citep{lee-2023, liu-2024} and machine translation, including idiom-aware evaluation \citep{Li_2024, khoshtab-2025}. 
However, these methods typically make use of minimal prompting, supplying only the idiom without its meaning.
Based on our findings, we hypothesize that model effectiveness can be substantially elevated by incorporating additional cues such as explicit meanings, contextual examples, or reasoning scaffolds.
To illustrate this point, we present case studies demonstrating that providing an idiom’s intended meaning leads to measurable gains in performance.
More related details are presented in Appendix~\ref{appendix_case_study}.

\begin{table}[t]
  \centering
  \small
  \setlength{\tabcolsep}{3pt}
  \begin{tabular}{l c c c c c}
    \toprule
    \textbf{Task} & \textbf{Meaning} & \textbf{{Kendall’s $\tau$}} & \textbf{{Spearman}} & \textbf{{Pearson}} \\
    \midrule
    \multirow{2}{*}{\makecell[l]{Sentence\\Generation}}
    & \cellcolor{yellow!20}\cmark & \cellcolor{yellow!20}0.739 & \cellcolor{yellow!20}0.781 & \cellcolor{yellow!20}0.784\\
    & \xmark & 0.118 & 0.125 & 0.110\\
    \midrule
    \multirow{2}{*}{\makecell[l]{Machine\\Translation}}
    & \cellcolor{yellow!20}\cmark & \cellcolor{yellow!20}0.615 & \cellcolor{yellow!20}0.672 & \cellcolor{yellow!20}0.671\\
    & \xmark & 0.403 & 0.449 & 0.453\\
    \bottomrule
  \end{tabular}
  \caption{Alignment between human evaluation and GPT-4o evaluation under two settings: with idiom meaning provided (\cmark) and without it (\xmark). Since GPT-4o achieves better automatic evaluation performance (i.e, aligns more closely with human judgments) when given the idiom meaning, we include the meaning information during model-based evaluation in Figure \ref{fig:case1}.}
  \label{tab:correlation}
\end{table}

\subsection{LLM-as-a-Judge on Idiom-Related Tasks}

We begin by cautioning against the na\"ive reliance on LLMs for automatic evaluation of idiom-related tasks, as is commonly seen in previous studies \cite{Li_2024}.
Our findings in \S\ref{subsec:context} provide strong evidence that supplying extra meaning information can improve LLM performance on idiom-related tasks---even when the models function as evaluators.
To verify this, we compare GPT-4o’s performance as a judge under two settings---one with the idiom’s meaning provided and one without. 
For each idiom-containing sentence from two tasks (detailed in the next subsection), the model assigns a 1–3 score based on how well the sentence conveys the idiom’s intended meaning.
We evaluate the effectiveness of the GPT-4o judge by examining its correlation with human-annotated scores.

Table~\ref{tab:correlation} reports that the GPT-4o judge aligns more closely with human annotations when given meaning information, indicating that conventional prompting fails to fully leverage LLMs for evaluating idiom-related tasks.
We therefore highlight MIDAS’s role in providing such information and adopt the meaning-augmented judge in subsequent experiments.


\begin{figure}[t]
  \centering
  \includegraphics[width=0.85\linewidth]{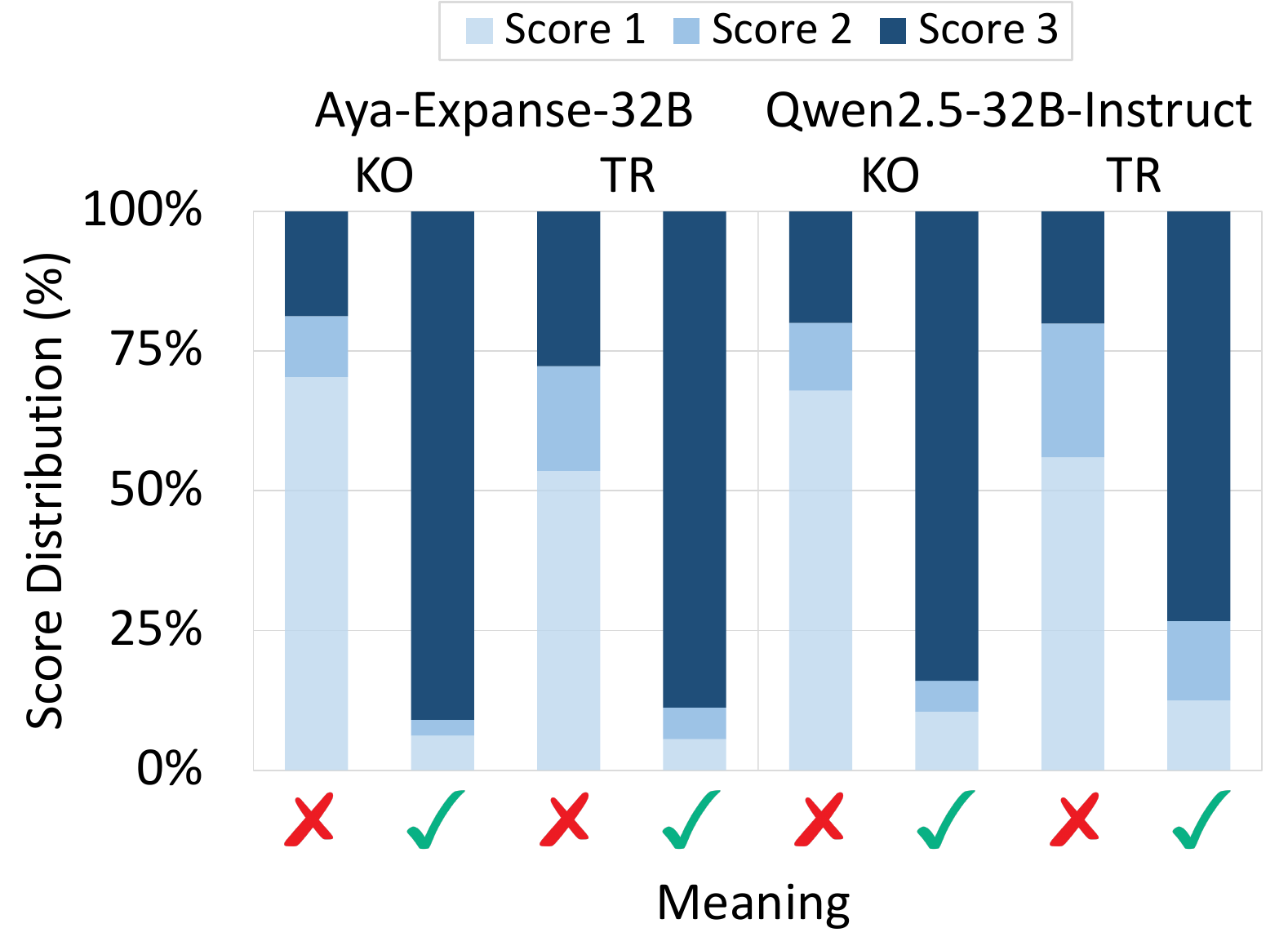}
  \caption{Sentence generation scores (1–3), rated by the meaning-augmented GPT-4o. Aya and Qwen are evaluated with (\cmark) and without (\xmark) idiom meanings. Meaning information boosts performance.}
  \label{fig:case1}
\end{figure}

\subsection{Idiom-Related Downstream Tasks}

This section reaffirms the utility of MIDAS-provided meaning information by analyzing its effect on idiom-related downstream tasks.
We focus on two tasks---\textbf{sentence generation} and \textbf{translation}---using two models (Qwen2.5-32B-Instruct and Aya-Expanse-32B) and two relatively low-resource languages (Korean and Turkish).
The task-specific details are as follows:
\begin{itemize}[leftmargin=10pt]
    \item \textbf{Sentence generation:} We compare models’ ability to generate idiom usage examples with and without access to the idiom’s meaning. 
    \item \textbf{Translation:} We translate to English under two settings: using either a GPT-4o-generated meaning or one from MIDAS.
\end{itemize}
In addition, we consider only idioms that (1) include at least one example sentence in MIDAS and (2) are consistently answered incorrectly across all three trials of the idiom MCQ task, indicating insufficient idiom understanding.

Figure~\ref{fig:case1} shows that in both languages, LLMs generate sentences that more accurately capture an idiom’s sense when its meaning is provided. 
A similar trend is observed in translation (Figure~\ref{fig:case2}), where models consistently perform better with access to meaning information. 
These results underscore the importance of applying appropriate strategies for idiom processing with LLMs. 
Illustrative examples are provided in Appendix~\ref{case:ex}.



\section{Conclusion}
In this work, we introduce MIDAS, a large-scale multilingual idiom dataset used to evaluate the idiom understanding capabilities of LLMs across various factors. 
Our findings show that LLMs adopt a hybrid approach, combining memorization with compositional reasoning, and are sensitive to contextual cues---although the effect of reasoning remains inconsistent. 
In future work, we plan to develop methods for improved idiom processing.

\section*{Limitations}
We outline several directions for future research to build upon this work.

\paragraph{Model size variability}
Our analysis focused on four distinct models with unique characteristics, but did not examine models of varying sizes within the same family. 
While this choice reflects our primary focus on examining whether the influence of various factors in idiom processing persists and generalizes across models with differing architectures and training regimes, it may be worthwhile to investigate the effects that arise solely from variations in model size.
We encourage future work to investigate how the influence of such factors varies with model size within the same model class.

\paragraph{Language coverage}
Although we cover six typologically and culturally diverse languages, this remains insufficient to capture the full diversity of the world's languages, each with its own set of unique idioms reflecting its culture. 
Future work should expand the language set to include understudied and endangered languages. 

\paragraph{Measurement of compositionality}
The method of measuring compositionality in this work relies on the models themselves. 
While we believe this offers a reasonable and scalable way to approximate idiom compositionality across languages, a more systematic approach (one that decouples the evaluation from the models being tested) would strengthen the analysis. 
To address this, future work could focus on developing methods for assigning compositionality scores without relying on LLMs.

\section*{Acknowledgments}
This work was supported by the National Science and Technology Major Project (Project 2022ZD0116306).
This work was supported by Institute of Information \& communications Technology Planning \& Evaluation (IITP) grant funded by the Korea government(MSIT) (No.RS-2020-II201373, Artificial Intelligence Graduate School Program(Hanyang University)).
This work was supported by the National Research Foundation of Korea(NRF) grant funded by the Korea government(MSIT) (RS-2025-00558151).
This research was supported by the Basic Science Research Program through the 
National Research Foundation of Korea (NRF) funded by the Ministry of Education (RS-2024-00353563).

\bibliography{anthology,custom}

\clearpage
\appendix
\section{Appendix: Dataset and MCQ Details}\label{appendix_A}

\subsection{Dataset Statistics}\label{appendix_A.1}
\begin{figure}[t!]
  \centering
  \includegraphics[width=0.9\linewidth]{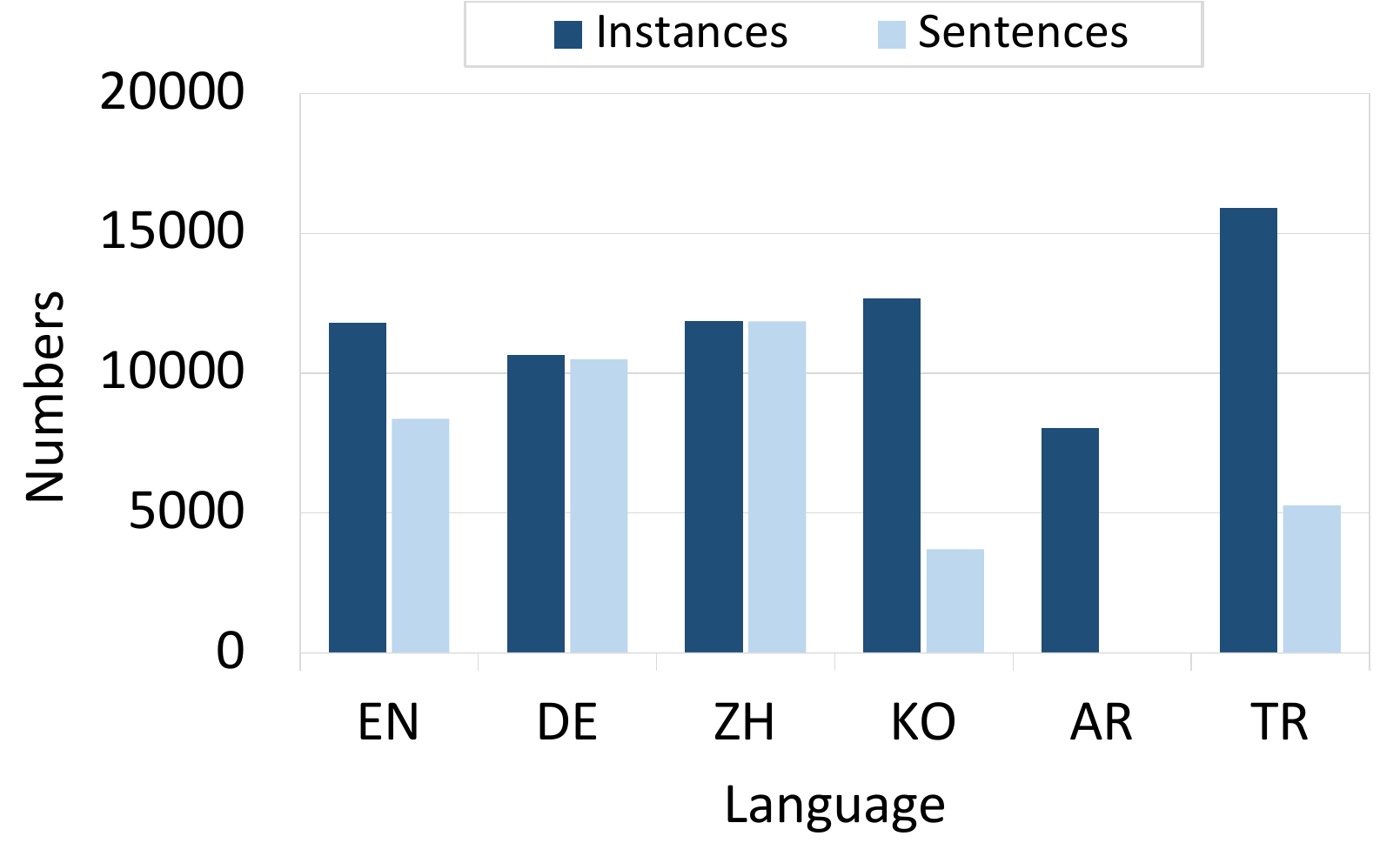}
  \caption{Statistics of our multilingual idiom dataset.}
  \label{fig:stat}
\end{figure}

\begin{figure}[t!]
  \centering
  \includegraphics[width=0.6\linewidth]{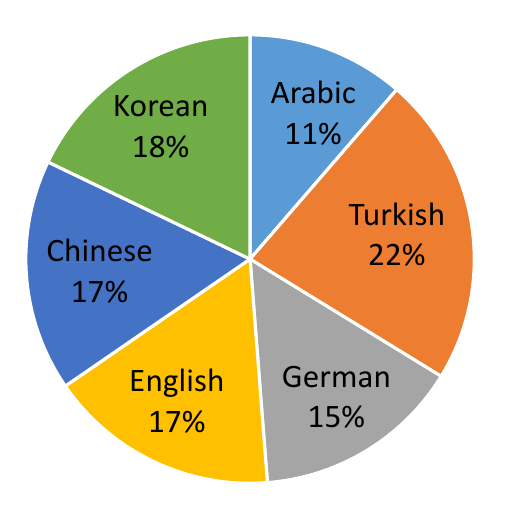}
  \caption{Language Distribution in our multilingual idiom dataset (n=70{,}909).}
  \label{fig:pie}
\end{figure}

Table \ref{tab:stat} and Figure \ref{fig:stat} report the exact counts and visualize the number of idiom instances and usage example sentences for each language in our corpus. Figure \ref{fig:pie} illustrates the language distribution in our multilingual idiom dataset.
Overall, the dataset consists of 70,909 instances covering 64,660 unique idiomatic expressions, with 39,696 instances containing at least one example sentence. Chinese (ZH) achieves full example coverage, while Arabic (AR) has no sentences available due to source constraints. 

\begin{table}[ht]
  \centering
  \small
  \begin{tabular}{l r r r}
    \toprule
    \textbf{Language} & \textbf{Uniq. Idioms} & \textbf{Instances} & \textbf{w/ Examples} \\
    \midrule
    EN  & 9,766  & 11,806 & 8,367 \\  
    DE   & 10,097 & 10,642 & 10,493 \\  
    ZH  & 11,851 & 11,851 & 11,851 \\
    KO   & 11,316 & 12,673 & 3,706 \\  
    AR   & 8,051  & 8,051  & 0 \\  
    TR  & 13,579 & 15,886 & 5,279 \\  
    \bottomrule
  \end{tabular}
  \caption{
    Statistics of our multilingual idiom dataset. 
    Note that the number of instances can exceed the number of unique idioms, as idioms with identical surface forms but different meanings are represented as separate entries.
  }
  \label{tab:stat}
\end{table}

\subsection{Dataset Example}\label{appendix_A.2}
\begin{figure*}[t!]
  \centering
  \includegraphics[width=0.8\linewidth]{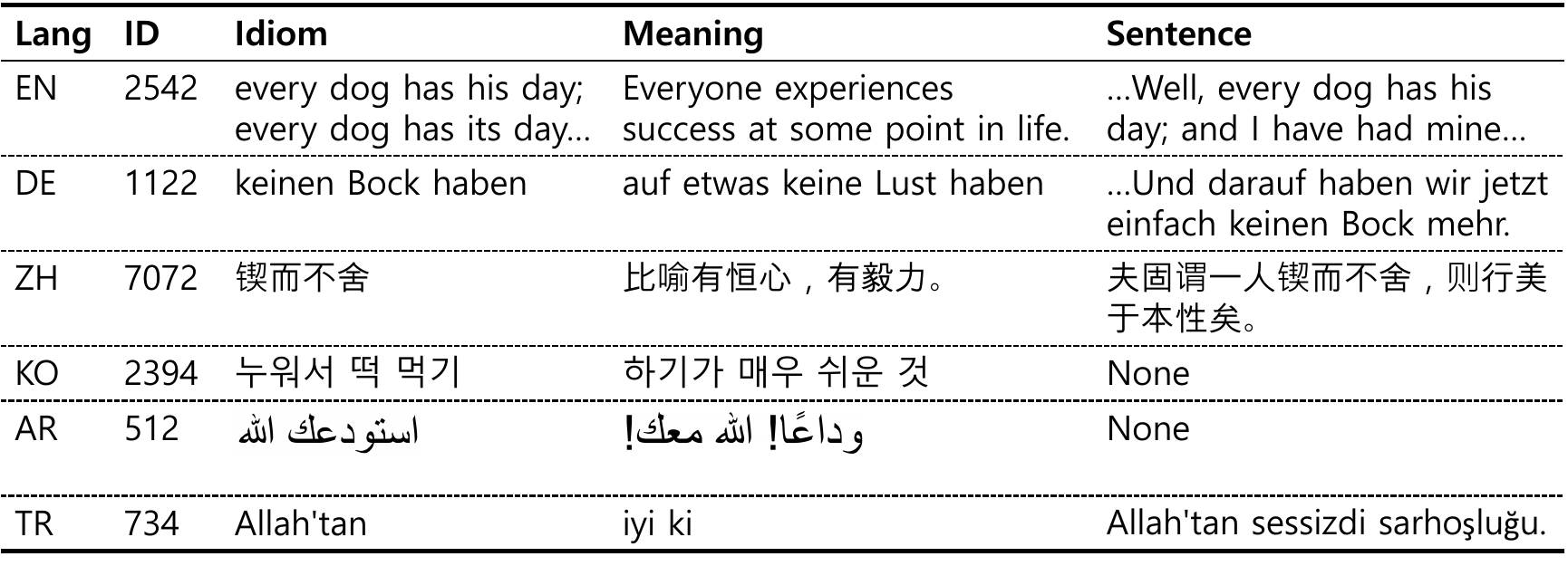}
  \caption{Example idiom entries for each language. Each row consists of an ID, a list of all possible idiom form variations, meaning, and sentence (if available).}
  \label{fig:data_eg}
\end{figure*}

Figure~\ref{fig:data_eg} gives one representative entry per language subset. Each idiom instance is given a unique ID, and if the idiom has many potential form variations (e.g. every dog has his/its day), the variations are stored within a single list. If a single idiom expression is associated with multiple distinct senses, we contain them in separate rows and append “-1”, “-2” to the ID to distinguish them.

\subsection{Dataset Construction}\label{appendix_A.3}
Our dataset construction follows a sequence of processes that can be summarized as (1) extraction from sources such as e-books, (2) basic preprocessing, and (3) meaning refinement.
After going through the initial steps, we serialize each language subset as JSON files following a uniform schema: \textit{ID}, \textit{Idiom} (list of surface variants), \textit{Meaning}, and \textit{Sentence} (list of usage example sentences). Further details on meaning refinement and language-specific considerations are described below:

\paragraph{Meaning refinement} As noted in \S\ref{sec:midas}, idiom meanings in our dataset are refined using LLMs prompted with the meaning of the target idiom and instructed to extract and retain only its core semantic content. While the LLMs are informed that the input represents the figurative meaning of an anonymous idiom, the idiom expression itself is not provided, as doing so could introduce bias. 

The primary goal of this refinement process is to remove extraneous details (such as explicit explanations of its figurative mechanism or origin) that could compromise the usefulness of the meanings as gold labels for subsequent automated MCQ construction and evaluation.

For example, consider the Korean idiom `가까운 데를 가도 점심밥을 싸 가지고 가거라' (literally, \textit{even if you are going somewhere nearby, take your lunch with you}). Its meaning is originally recorded as: `십 리밖에 안 되는 가까운 데를 가더라도 점심밥을 싸 가지고 다닌다는 뜻으로, \textbf{무슨 일에나 준비를 든든히 할 것}을 비유적으로 이르는 말,' which translates to: \textit{An expression meaning that even when going to a nearby place, one takes a packed lunch—used figuratively to \textbf{advise thorough preparation for any undertaking}}. 
After refinement, this lengthy entry is reduced to its core semantic content: `\textbf{무슨 일에나 준비를 든든히 할 것} (\textbf{advises thorough preparation for any undertaking}).'

\paragraph{English(EN)} The English idioms included in our data are all extracted from documents categorized as ``English idioms" and ``English proverbs" in Wiktionary\footnote{\url{https://en.wiktionary.org/wiki/Category:English\_idioms}} using Beautiful Soup\footnote{\url{https://pypi.org/project/beautifulsoup4/}}, yielding 9,766 idioms. 
In the preprocessing step, we run custom Python scripts to correct common parsing errors, normalize whitespace and punctuation, and expand certain abbreviations (e.g.\ converting “sth” to “something”). 
All meanings are first refined with GPT-4o mini, after which a human annotator with a background in English linguistics verified the form and meaning.

\paragraph{German(DE)} The German idioms in our dataset are primarily extracted from the e-book version of \textit{Duden – Redewendungen}.\footnote{\url{https://shop.duden.de/Duden-Redewendungen/9783411041152}}
A total of 10,097 idioms are obtained using the PyMuPDF library.\footnote{\url{https://github.com/pymupdf/PyMuPDF}} 
This library is used to extract the idiom expressions, their meanings, and example sentences based on font characteristics, text size, and x, and y-axis coordinates. 
In preprocessing, we remove irrelevant content, correct German character encoding errors (e.g.\ garbled “ä”, “ü”), normalize whitespace and punctuation, merge hyphenated fragments, and expand abbreviations (e.g.\ “etw” to “etwas”) via automation where possible; any items that could not be processed automatically are manually handled by a native German annotator. 
All meanings are then refined with GPT-4o, after which the annotator performed a final pass of semantic validation.

\paragraph{Chinese(ZH)} We use the \texttt{idiom.json} file from the \textit{chinese-xinhua} repository\footnote{\url{https://github.com/pwxcoo/chinese-xinhua}}, which provides structured idiom entries sourced from the Xinhua Dictionary. To maintain language balance across our dataset, we select 11,851 idioms that include example sentences. Definitions are refined using GPT-4o with a prompt designed to retain only the essential meaning—excluding component character breakdowns, synonymous idioms, or etymological details—while preserving the original phrasing.

\paragraph{Korean(KO)} To construct the Korean dataset, we begin by extracting 11,316 Korean idioms from the \textit{Korean Standard Dictionary}\footnote{\url{https://stdict.korean.go.kr/main/main.do}} provided by the National Institute of Korean Language. Meanings are refined with GPT-4o to consolidate overly specific senses. A native Korean annotator then performs preprocessing—correcting spacing errors, merging variant forms, and standardizing punctuation—followed by a final semantic review to ensure fidelity and internal consistency.

\paragraph{Arabic(AR)} We build our Arabic subset upon \textit{A Dictionary of Arabic Idioms and Expressions} by the Edinburgh University Press,\footnote{\url{https://edinburghuniversitypress.com}} we obtain 8,051 idioms (the source provides no usage examples) via the PyMuPDF\footnote{\url{https://github.com/pymupdf/PyMuPDF}} library. This library is used to extract the idiom expressions, their meanings, and example sentences based on font characteristics, text size, and x, and y-axis coordinates. Since the original dataset only provided meanings in English, we use GPT-4o to translate these into Arabic, and the translated meanings are further refined and validated by native Arabic annotator. Given the prevalence of parsing errors in right-to-left scripts (e.g., broken ligatures, misplaced diacritics), the annotator performs targeted preprocessing—character-shape normalization, word-boundary correction—before completing the final linguistic verification.

\paragraph{Turkish(TR)} Our Turkish dataset builds on the \textit{Turkish Idioms and Proverbs}\footnote{\url{https://www.kaggle.com/datasets/emreokcular/turkish-idioms-and-proverbs}} dataset from Kaggle, which was originally compiled using data from TDK (Türk Dil Kurumu, or Turkish Language Association)\footnote{\url{https://tdk.gov.tr/}}—yielding 13,579 idioms. For idiom instances with example sentences, the original data do not include separate columns for the sentences; instead, they are embedded within the meaning texts, which have to be separated. The meanings are then refined by a native Turkish annotator to ensure both semantic fidelity and stylistic consistency.

\subsection{Legal and Ethical Considerations}\label{appendix_A.4}
For the German, Arabic, and Turkish subsets, native-speaking student annotators were recruited to perform data preprocessing and quality validation. The German annotator, a computer-science major, and the Arabic annotator, a data-science major, were selected to ensure adequate computational proficiency. Each annotator received a comprehensive PDF guide outlining the step-by-step pipeline for preprocessing our dataset and performing quality verification. The German subset required 9 hours, the Turkish 5 hours, and the Arabic 11 hours to complete; compensation was set at approximately 1.5 times the legally mandated minimum hourly wage.
To ensure strict compliance with copyright and licensing requirements, every dataset was procured exclusively through legally sanctioned sources or under open licenses. Additional licensing details are presented below. 

\paragraph{English(EN)} The English subset is licensed under the Creative Commons Attribution–ShareAlike 4.0 International (CC BY-SA 4.0)\footnote{\url{https://creativecommons.org/licenses/by-sa/4.0/}}. Accordingly, all entries are attributed to the Wiktionary and distributed under the same terms. Any modifications are clearly indicated, and derivative works are shared under the identical license.

\paragraph{German(DE)} We contacted Duden\footnote{\url{https://www.duden.de/}} via email to request permission for public use of the dataset but did not receive a response. The dataset will be made publicly available upon approval. Until then, the German subset remains private.

\paragraph{Chinese(ZH)} The Chinese idioms are sourced from the \textit{chinese-xinhua} GitHub repository, which compiles content from \texttt{zdic.net}. According to the site's copyright policy,\footnote{\url{https://www.zdic.net/aboutus/copyright/}} 
all dictionary materials are released under the CC0 1.0 Public Domain Dedication. 

\paragraph{Korean(KO)} The Korean dataset is licensed under the Creative Commons Attribution–ShareAlike 2.0 Korea (CC BY-SA 2.0 KR)\footnote{\url{https://creativecommons.org/licenses/by-sa/2.0/kr/}}. Accordingly, all entries in the Korean dataset are attributed to the Standard Korean Language Dictionary (National Institute of Korean Language) and distributed under the same terms. Any modifications are clearly indicated, and derivative works are shared under the identical license.

\paragraph{Arabic(AR)} Email correspondence with the publisher confirmed that reuse of the dataset is permitted under fair-dealing provisions.

\paragraph{Turkish(TR)} Email correspondence with the original publisher (TDK) indicates that permission to publicly share the dataset could not be granted; consequently, the Turkish subset remains private.

\subsection{MCQ Construction}
Algorithm \ref{alg:mcq_pseudocode} outlines our MCQ construction procedure. 
For each language subset, we employ the \textit{intfloat/multilingual-e5-large-instruct} model to compute similarity scores between (1) idiom expressions and all meanings in the same language, and (2) meanings and all other meanings in the same set. 
For each idiom, we discard the top 1\% of highly similar candidates from both groups to avoid multiple valid answers. 
We then sample two distractors from each group, producing MCQs with four distractors and one correct answer corresponding to the idiom’s true meaning. 
During evaluation, the options are shuffled so that the correct answer appears in three different positions across trials.

\begin{algorithm}[t]
\small
\caption{\small MCQ construction with shuffled answer choices}
\begin{spacing}{1.15} 
\begin{algorithmic}[1]
\Require A dataset $\mathcal{D}_{\text{MIDAS}} := \{i_j:=(s_j, m_j)\}_{j=1}^N$, where $i$: an idiom, $s$: its surface form, $m$: its meaning
\Ensure An MCQ with three shuffled sets of answer choices
\State Encode: $\forall j, \mathbf{e}_{s_j}\gets\text{Embed}(s_j), \mathbf{e}_{m_j}\gets\text{Embed}(m_j) $
\State Sample: $k \sim \{1,\dots,N\}$
\State Compute two similarity-based score sets centered on $i_k$:
\Statex $\sigma_{\text{surf}} := \{\cos(\mathbf{e}_{s_k},\mathbf{e}_{m_j})|\forall j \in \{1,\dots,N\},j \ne k\}$,
\Statex $\sigma_{\text{mean}} := \{\cos(\mathbf{e}_{m_k},\mathbf{e}_{m_j})|\forall j \in \{1,\dots,N\}, j \ne k\}$ 
\State Sort $\sigma_{\text{surf}}$ and $\sigma_{\text{mean}}$ in descending order
\State Remove top 1\% items from $\sigma_{\text{surf}}$ and $\sigma_{\text{mean}}$
\State $\mathcal{D}_{\text{surf}}$ and $\mathcal{D}_{\text{mean}}$ $\leftarrow$ retrieve top-2 items from $\sigma_{\text{surf}}$ and $\sigma_{\text{mean}}$, and extract their corresponding meanings ($m_j$)
\State Let $\mathcal{O} := \{ m_k \} \cup \mathcal{D}_{\text{surf}} \cup \mathcal{D}_{\text{mean}}$
\State Initialize $\mathcal{Q} \gets \phi$ (empty set)
\For{$t = 1$ to $3$}
    \Repeat
        \State $O^{(t)} \leftarrow$ Shuffle($\mathcal{O}$)
    \Until{$\text{index}(m_k \in O^{(t)}) \ne \text{index}(m_k \in Q), \forall Q \in \mathcal{Q}$}
    \State $\mathcal{Q} \leftarrow \mathcal{Q} \cup \{O^{(t)}\}$
\EndFor
\State \Return $(s_k, \mathcal{Q})$
\end{algorithmic}
\end{spacing}
\label{alg:mcq_pseudocode}
\end{algorithm}

\section{Appendix: Memorization Details}

\subsection{Continuation Candidate Filter}\label{appendix_continuation_filter}
\begin{itemize}[leftmargin=8pt]
    \item \textbf{Idiom length:} To account for linguistic differences (E.g., isolating languages such as English and agglutinative languages such as Korean), we apply language-specific thresholds: filtering out idioms with fewer than four words in English, German, and Chinese, and fewer than three in Korean, Arabic, and Turkish.
    \item \textbf{Context-target similarity:} We use FastText \cite{bojanowski-2017} models of each language to compare the embeddings of context tokens with those of the target token. Idioms are excluded if any context token has a cosine similarity above 0.7 with the target.\footnote{We tested a small set of idioms in multiple languages to determine a threshold that generalizes across all languages.}
    \item \textbf{Subsequence predictability:} For each model, we exclude idiom instances where the target word can be predicted from only 1–4 preceding tokens, rather than full context---that is, if the model can predict the target word $w_n$ using subsequences such as $w_{n-1}, \dots, w_{n-4}$ within the idiom, the instance is removed from the candidate set.
    
    Formally, we exclude the idiom if
\begin{align*}
\quad 
& \arg\max_{w} P(w \mid w_{1}, \ldots, w_{n-1}) \\
& = \arg\max_{w} P(w \mid w_{n-k}, \ldots, w_{n-1}) \\
& = w_n (\exists k \in \{1, 2, 3, 4\})
\end{align*}
    \item \textbf{Overlapping context:} Some idioms such as ``back in the \textit{day}" and ``back in the \textit{game}" share an overlapping context. We exclude such instances from our continuation candidates.
\end{itemize}

\subsection{Continuation Candidate Statistics}\label{appendix_continuation_candidate}
Table~\ref{tab:our-dataset} shows the number of instances before and after applying our continuation candidate filtering. Note that the number of candidates varies by model, as each is subject to the third condition mentioned above.

\begin{table}[ht]
  \centering
  \scriptsize
  \setlength{\tabcolsep}{4pt}
  \begin{tabular}{l r r r r r}
    \toprule
    \textbf{Lang.} & \textbf{Original} & \textbf{Aya-Expanse} & \textbf{Qwen2.5} & \textbf{DeepSeek-V3} & \textbf{GPT-4o} \\
    \midrule
    EN  &  9{,}766 & 2,336 & 2,546 & 2,328 & 2,114 \\  
    DE  & 10{,}097 & 5,440 & 6,173 & 5,250 & 4,853 \\  
    ZH  & 11{,}851 & 8,200 & 7,935 & 6,495 & 6,787 \\
    KO  & 11{,}316 & 6,943 & 7,143 & 6,935 & 6,698 \\  
    AR  &  8{,}051 & 2,780 & 2,733 & 2,815 & 2,708 \\  
    TR  & 13{,}579 & 6,978 & 7,267 & 6,905 & 6,689 \\  
    \bottomrule
  \end{tabular}
  \caption{Number of idiom instances before and after applying our continuation candidate filter.}
  \label{tab:our-dataset}
\end{table}

\subsection{Memorization Sample Statistics}\label{appendix_memorization_sample}
The number of candidates and memorization rates varied across models and languages, resulting in imbalanced sizes between memorized and unmemorized groups. To mitigate potential bias from this imbalance, we perform per-model sampling based on the smallest group, ensuring equal numbers of instances in the memorized and unmemorized sets for each language in our analysis of the correlation between idiom understanding and memorization (\S\ref{subsec:memorization}). For each model, we sample the same number of idioms from both groups—that is, 459 memorized and 459 unmemorized idioms for Aya-Expanse, 682 each for Qwen2.5, 681 each for DeepSeek-v3, and 731 each for GPT-4o.

\subsection{Examples of Unmemorized Idioms}\label{appendix_unmemorized_examples}
\begin{figure}[t]
  \centering
  \includegraphics[width=\linewidth]{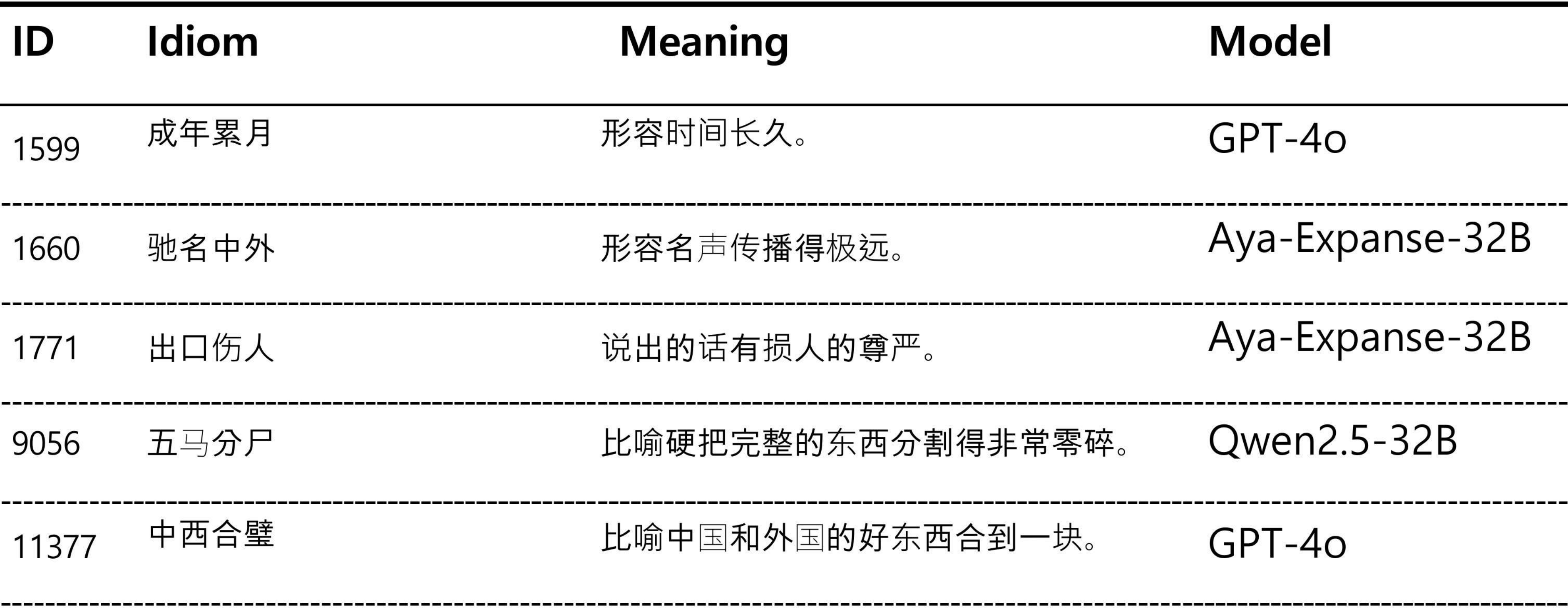}
  \caption{Examples of unmemorized Chinese idioms. The \textbf{Model} column specifies which model failed to memorize each idiom.}
  \label{fig:unmemorized}
\end{figure}

While most unmemorized idioms are relatively uncommon, we also observe notable exceptions, such as the Chinese idioms shown in Figure \ref{fig:unmemorized}.
This is particularly intriguing given the characteristics of Chinese idioms, whose predictable semantics and rigid four-character structure ensure they consistently appear in the same form across different contexts.
However, we believe this observation does not contradict the connection between an idiom’s social recognition and its memorization; rather, it highlights the complexity underlying idiom memorization in LLMs.

Although it is likely that highly recognized idioms are used more frequently, it is unclear whether the models we discuss are trained on data that represent such characteristics properly. Furthermore, there is the possibility of various other aspects that might influence idiom memorization, such as compositionality, as shown in \ref{appendix_factor}. 
What these aspects are, and the extent to which they influence idiom memorization, remain unclear and warrant detailed investigation beyond the scope of this study. We therefore leave such questions to future work.

\section{Appendix: Compositionality Details}\label{appendix_Comp}

\subsection{Compositionality Score Validation on Noun Compounds}\label{appendix_NC}
To evaluate the validity of our compositionality scoring method, we also apply it to non-idiomatic expressions. Specifically, we conduct a small-scale experiment using 111 fully compositional English noun compounds (e.g., police court → “court for minor offenders arrested by the police”) drawn from the test set of \citealp{coil-2023}. These expressions were evaluated using the same LLM-based framework employed for idioms. Table \ref{tab:NC} presents the average compositionality scores for both idioms and noun compounds, which range from 1 to 5, with higher scores indicating greater compositionality. As expected, models consistently assigned high compositionality scores to these fully compositional expressions, in contrast to the more varied ratings observed for idioms. These results further support the validity of our LLM-based approach.

\begin{table}[ht]
    \centering
    \small
    \begin{tabular}{lcc}
    \toprule
    \textbf{Models} & \textbf{EN Idioms} & \textbf{EN Noun Compounds} \\
    \midrule
    Aya-Expanse      & 2.956 & 4.582 \\
    Qwen2.5          & 2.340 & 4.973 \\
    Deepseek-V3      & 2.863 & 4.736 \\
    GPT-4o           & 2.670 & 4.927 \\
    \bottomrule
    \end{tabular}
    \caption{Comparison of compositionality scores of models on EN idioms from MIDAS and EN noun compounds from \citealp{coil-2023}. A significant gap emerges between the two groups, reflecting the models’ ability to capture varying levels of compositionality.}
    \label{tab:NC}
\end{table}

\subsection{Compositionality Score Statistics}\label{appendix_CS_Stat}
We report the average compositionality scores of all models and languages in Table \ref{tab:avg_comp}.

\begin{table}[ht]
    \centering
    \scriptsize
    \setlength{\tabcolsep}{4pt}
    \begin{tabular}{lcccc}
    \toprule
    \textbf{Language} & \textbf{Aya-Expanse} & \textbf{Qwen2.5} & \textbf{Deepseek-V3} & \textbf{GPT-4o} \\
    \midrule
    EN & 2.956 (0.514) & 2.340 (0.937) & 2.863 (0.868) & 2.670 (0.887) \\
    DE & 3.110 (0.375) & 1.834 (0.785) & 2.590 (0.690) & 2.398 (0.825) \\
    ZH & 3.130 (0.490) & 2.295 (0.981) & 2.925 (0.802) & 2.193 (1.032) \\
    KO & 2.966 (0.334) & 1.382 (0.576) & 2.222 (0.744) & 1.853 (0.679) \\
    AR & 3.466 (0.589) & 2.351 (1.162) & 2.651 (0.829) & 3.107 (1.233) \\
    TR & 3.098 (0.378) & 1.658 (0.787) & 2.421 (0.707) & 2.176 (0.862) \\
    \bottomrule
    \end{tabular}
    \caption{Mean compositionality scores by language and model with standard deviations in parentheses.}
    \label{tab:avg_comp}
\end{table}

\section{Appendix: Prompt Details}\label{appendix_Prompt}
\subsection{Zero-shot MCQ}\label{appendix_ZeroShot}
This section presents the prompt formats used for the base MCQ task. Prompts consist of a question asking for the idiomatic meaning of a given expression, followed by five answer choices. We provide two examples of our prompt, the English version in  Figure~\ref{fig:EN-MCQ} and the Korean version in Figure~\ref{fig:KO-MCQ}.

\begin{figure}[ht]
  \setlength{\fboxsep}{7pt}
  \noindent
  \begin{tcolorbox}[
      colback=gray!10,       
      colframe=gray!75,      
      sharp corners=south,   
      rounded corners=northwest,
      boxrule=0.8pt,         
      width=\columnwidth,    
      fonttitle=\bfseries,
      coltitle=black,        
      title=EN MCQ prompt
  ]
    \fontsize{9pt}{11pt}\selectfont
    \setlength{\parskip}{5pt}

    What is the idiomatic meaning of the idiom \texttt{\{idiom\}}? Choose from the options below.

    \begin{enumerate}[label=\arabic*.]
      \item \texttt{\{option 1\}}
      \item \texttt{\{option 2\}}
      \item \texttt{\{option 3\}}
      \item \texttt{\{option 4\}}
      \item \texttt{\{option 5\}}
    \end{enumerate}

    Respond with \textbf{ONLY} the number (1, 2, 3, 4, or 5). Do \textbf{NOT} add any extra text, punctuation, or explanation.
  \end{tcolorbox}
  \caption{English version of our zero-shot MCQ prompt.}
  \label{fig:EN-MCQ}
\end{figure}

\begin{figure}[ht]
  \setlength{\fboxsep}{7pt}
  \noindent
  \begin{tcolorbox}[
      colback=gray!10,       
      colframe=gray!75,      
      sharp corners=south,   
      rounded corners=northwest,
      boxrule=0.8pt,         
      width=\columnwidth,    
      fonttitle=\bfseries,
      coltitle=black,        
      title=KO MCQ prompt
  ]
    \fontsize{9pt}{11pt}\selectfont
    \setlength{\parskip}{5pt}

    관용 표현 \texttt{\{idiom\}}의 관용적 의미는 무엇인가요? 아래 보기 중에서 가장 알맞은 것을 고르세요.

    \begin{enumerate}[label=\arabic*.]
      \item \texttt{\{option 1\}}
      \item \texttt{\{option 2\}}
      \item \texttt{\{option 3\}}
      \item \texttt{\{option 4\}}
      \item \texttt{\{option 5\}}
    \end{enumerate}

    반드시 숫자 (1, 2, 3, 4, 또는 5)만 입력하세요. 추가 설명이나 기호는 쓰지 마세요.
  \end{tcolorbox}
    \caption{Korean version of our zero-shot MCQ prompt.}
    \label{fig:KO-MCQ}
\end{figure}

\subsubsection*{Other Languages (DE, ZH, AR, TR)}

Prompts in other languages follow the same structure as above, with all text translated appropriately. Each language also includes localized phrasing for the idiom definition and instructions to ensure cultural and linguistic clarity.

\subsection{Configurations for Context Experiments}
\label{appendix_ContextPrompt}
This section presents the prompt formats used for the MCQ task under the \textit{with-context} and \textit{without-context} conditions. The same zero-shot MCQ prompt is given in the \textit{without-context} condition, while in the \textit{with-context} condition, a sentence containing the idiom is additionally provided to the zero-shot MCQ prompt as in Figure \ref{Fig:context_prompt}.

\begin{figure}[ht]
  \setlength{\fboxsep}{7pt}
  \noindent
  \begin{tcolorbox}[
      colback=gray!10,       
      colframe=gray!75,      
      sharp corners=south,   
      rounded corners=northwest,
      boxrule=0.8pt,         
      width=\columnwidth,    
      fonttitle=\bfseries,
      coltitle=black,        
      title=With Context prompt
  ]
    \fontsize{9pt}{11pt}\selectfont
    \setlength{\parskip}{5pt}

    What is the idiomatic meaning of the idiom \texttt{\{idiom\}}? Choose from the options below.

    \textcolor{cyan}{Here is a sentence that includes the idiom: \texttt{\{sentence\}}}

    \begin{enumerate}[label=\arabic*.]
      \item \texttt{\{option 1\}}
      \item \texttt{\{option 2\}}
      \item \texttt{\{option 3\}}
      \item \texttt{\{option 4\}}
      \item \texttt{\{option 5\}}
    \end{enumerate}

    Respond with \textbf{ONLY} the number (1, 2, 3, 4, or 5). Do \textbf{NOT} add any extra text, punctuation, or explanation.
  \end{tcolorbox}
  \caption{Prompt for our \textbf{With Context} setting, where the \textcolor{cyan}{highlighted} part is added to our zero-shot MCQ prompt.}
  \label{Fig:context_prompt}
\end{figure}



\section{Appendix: Additional Results and Examples}\label{appendix_additional_results}

\subsection{Accuracy with and without Context across Models}\label{appendix_context}

Table~\ref{tab:context_results_all_models} reports the accuracy (\%) for each model and language, with and without additional context. As shown, context consistently improves performance across all models and languages. 

\begin{table}[ht]
\centering
\scriptsize
\setlength{\tabcolsep}{4pt}
\begin{tabular}{lccc}
\toprule
\textbf{Model} & \textbf{Language} & \textbf{w/o Context (\%)} & \textbf{w/ Context (\%)} \\
\midrule
\multirow{5}{*}{\texttt{Aya-Expanse-32B}} 
& EN & 82.43 & 92.77 \\
& DE & 72.48 & 87.16 \\
& ZH & 74.82 & 82.57 \\
& KO & 52.78 & 82.81 \\
& TR & 47.95 & 75.15 \\
\midrule
\multirow{5}{*}{\texttt{Qwen2.5-32B}} 
& EN & 84.86 & 93.85 \\
& DE & 73.83 & 90.02 \\
& ZH & 92.58 & 94.63 \\
& KO & 51.46 & 82.14 \\
& TR & 37.34 & 69.21 \\
\midrule
\multirow{5}{*}{\texttt{DeepSeek-V3}} 
& EN & 91.10 & 95.76 \\
& DE & 84.78 & 92.96 \\
& ZH & 95.44 & 95.87 \\
& KO & 57.93 & 83.73 \\
& TR & 63.06 & 88.45 \\
\midrule
\multirow{5}{*}{\texttt{GPT-4o}} 
& EN & 91.53 & 95.22 \\
& DE & 88.29 & 94.63 \\
& ZH & 90.61 & 93.34 \\
& KO & 75.93 & 92.42 \\
& TR & 71.24 & 90.72 \\
\bottomrule
\end{tabular}
\caption{Accuracy (\%) of each model across five languages, with and without context.}
\label{tab:context_results_all_models}
\end{table}

\begin{figure}[ht]
\centering
  \includegraphics[width=\linewidth]{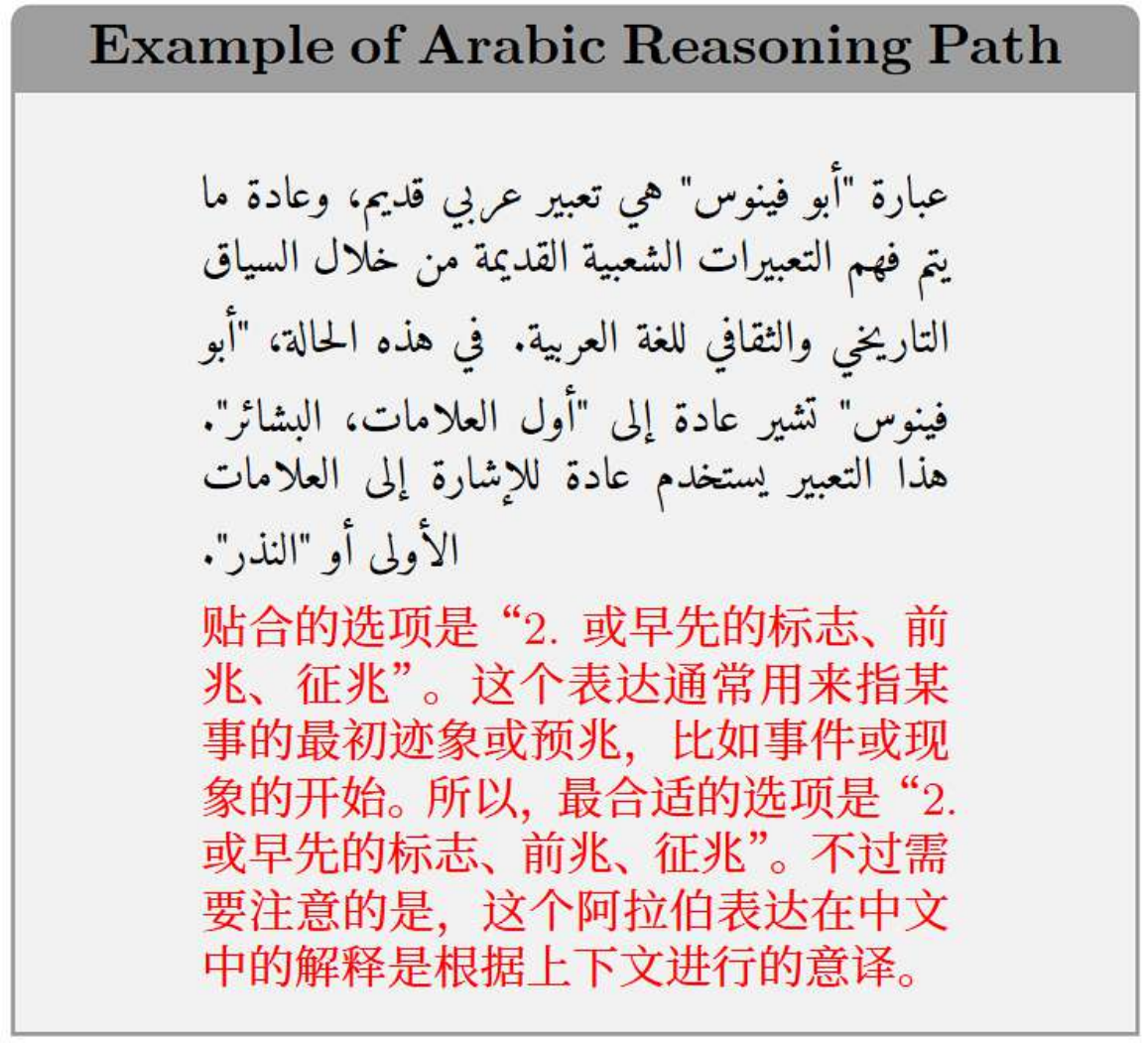}
  \caption{An example of Qwen2.5 generated reasoning path in Arabic including code-mixing in Chinese.}
  \label{fig:AR code-mixing}
\end{figure}

\subsection{Arabic Reasoning}\label{appendix_arabic}
We observe that when reasoning in Arabic, models often exhibit problematic behaviors (such as code-mixing or refusal to answer) that are not seen in other languages, frequently resulting in incorrect outputs.
Figure \ref{fig:AR code-mixing} presents an example of a reasoning path generated by Qwen that exhibits code-mixing and results in an incorrect answer.


\subsection{Additional Reasoning Results}\label{appendix_additional_reasoning}
We report results of our reasoning experiments expanded to Qwen2.5-32B/DeepSeek-R1-Distill-Qwen-32B and Llama-3.3-70B-Instruct/DeepSeek-R1-Distill-Llama-70B pairs in Table \ref{tab:qwen_R1} and Table \ref{tab:llama_R1}, respectively. Note that Llama 3.3 officially claims support only for English and German among the six languages we examine. 

The results exhibit patterns similar to those reported in \S\ref{main:reasoning}. 
For EN and DE, CoT leads to degraded performance, whereas it yields improvements for KO and TR. 
In Arabic, we also observed similar proficiency issues with Llama3.3, resulting in reduced accuracy under local CoT. 
Regarding the reasoning variants, performance in both Thinking and NoThinking often lags behind that of No-CoT. This underscores a potential drawback of reasoning-enhanced post-training: models may experience catastrophic forgetting on tasks that receive less emphasis during reasoning-focused training.

\begin{table}[ht]
    \centering
    \scriptsize
    \begin{tabular}{lcccc}
    \toprule
    \textbf{Language} & \textbf{CoT} & \textbf{No CoT} & \textbf{Thinking} & \textbf{No Thinking} \\
    \midrule
    EN & 87.49 & 87.99 & 85.88 & 86.86 \\
    DE & 69.80 & 74.88 & 68.35 & 69.56 \\
    ZH & 92.54 & 92.46 & 89.48 & 89.89 \\
    KO & 56.66 & 53.77 & 51.57 & 45.28 \\
    AR & 48.45 & 69.77 & 70.73 & 64.87 \\
    TR & 52.52 & 47.54 & 40.46 & 33.31 \\
    \bottomrule
    \end{tabular}
    \caption{MCQ accuracy (\%) of Qwen2.5 and its reasoning variant with and without reasoning across languages.}
    \label{tab:qwen_R1}
\end{table}

\begin{table}[ht]
    \centering
    \scriptsize
    \begin{tabular}{lcccc}
    \toprule
    \textbf{Language} & \textbf{CoT} & \textbf{No CoT} & \textbf{Thinking} & \textbf{No Thinking} \\
    \midrule
    EN & 90.94 & 91.45 & 89.92 & 88.07 \\
    DE & 75.18 & 76.70 & 74.34 & 68.80 \\
    ZH & 75.59 & 73.36 & 68.97 & 63.65 \\
    KO & 48.06 & 49.93 & 50.14 & 39.88 \\
    AR & 46.73 & 64.73 & 66.01 & 47.42 \\
    TR & 59.71 & 54.99 & 54.88 & 42.47 \\
    \bottomrule
    \end{tabular}
    \caption{MCQ accuracy (\%) of Llama3.3 and its reasoning variant with and without reasoning across languages.}
    \label{tab:llama_R1}
\end{table}

\begin{figure}[ht]
  \centering
  \includegraphics[width=0.9\linewidth]{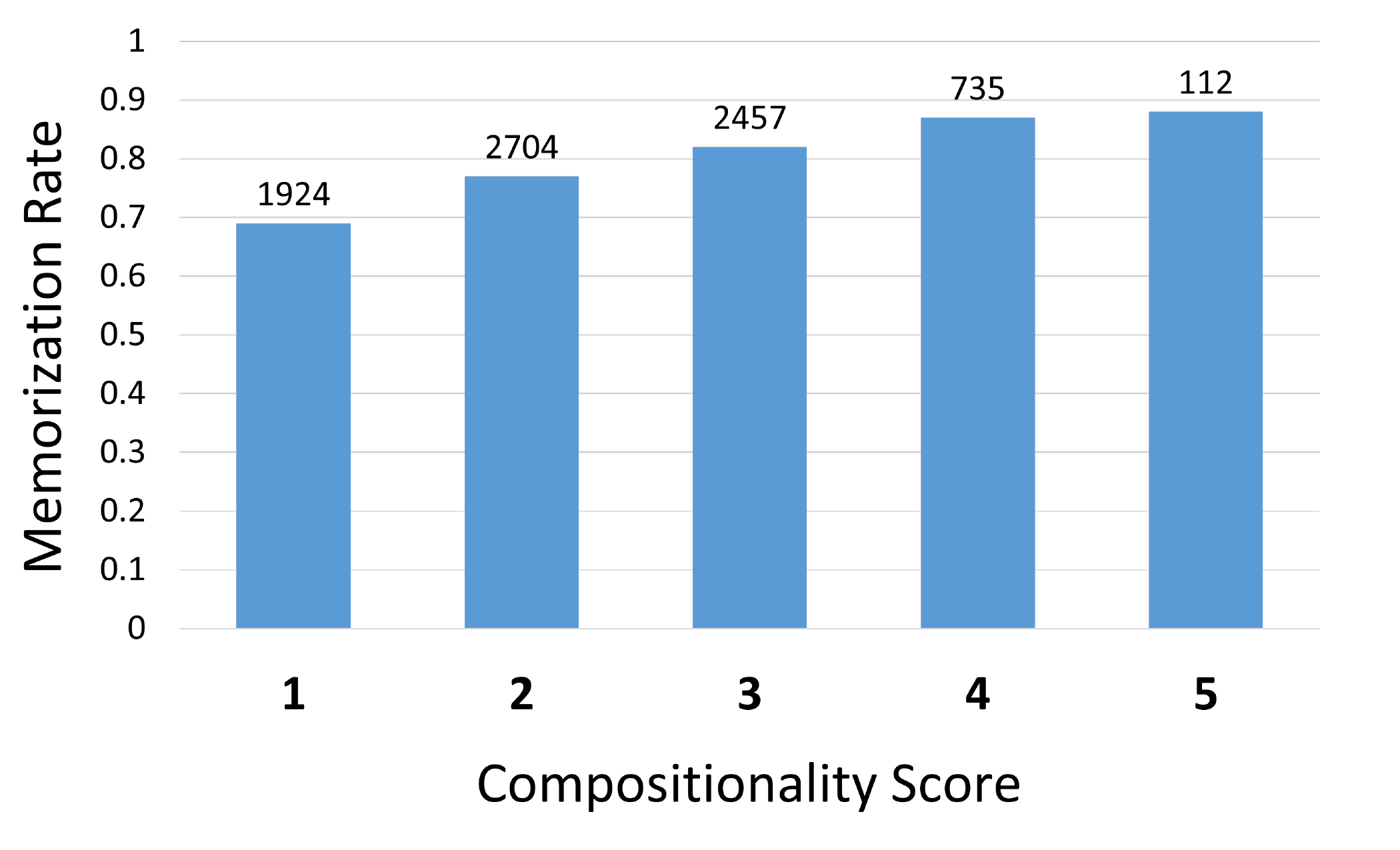}
  \caption{LLMs tend to memorize idiom instances with higher compositionality scores more easily, as observed in the Qwen2.5-32B-Instruct model for Chinese. The numbers above each bar indicate the number of idiom instances within each compositionality score bin.}
  \label{fig:mem-comp}
\end{figure}

\begin{table}[ht]
\centering
\scriptsize
\setlength{\tabcolsep}{4pt}
\begin{tabular}{llll}
\toprule
\textbf{Model} & \textbf{Language} & \textbf{Kendall's $\tau$} & \textbf{Spearman's $\rho$} \\
\midrule
Aya-Expanse-32B     & EN & 0.096 *** & 0.099 *** \\
        & DE & 0.120 *** & 0.121 *** \\
        & ZH & 0.039 *** & 0.040 *** \\
        & KO & 0.083 *** & 0.084 *** \\
        & AR & 0.138 *** & 0.141 *** \\
        & TR & 0.057 *** & 0.058 *** \\
\midrule
Qwen2.5-32B    & EN & 0.168 *** & 0.181 *** \\
        & DE & 0.122 *** & 0.129 *** \\
        & ZH & 0.130 *** & 0.141 *** \\
        & KO & 0.148 *** & 0.150 *** \\
        & AR & 0.128 *** & 0.140 *** \\
        & TR & 0.049 *** & 0.052 *** \\
\midrule
DeepSeek-V3 & EN & 0.153 *** & 0.164 *** \\
         & DE & 0.142 *** & 0.148 *** \\
         & ZH & 0.111 *** & 0.119 *** \\
         & KO & 0.146 *** & 0.154 *** \\
         & AR & 0.081 *** & 0.086 *** \\
         & TR & 0.055 *** & 0.058 *** \\
\midrule
GPT-4o  & EN & 0.131 *** & 0.140 *** \\
        & DE & 0.074 *** & 0.079 *** \\
        & ZH & 0.084 *** & 0.091 *** \\
        & KO & 0.101 *** & 0.106 *** \\
        & AR & 0.050 *   & 0.056 * \\
        & TR & -0.010    & -0.010 \\
\bottomrule
\end{tabular}
\caption{Kendall’s $\tau$ and Spearman’s $\rho$ correlations and their statistical significance (marked with $^*$ ($p{<}.05$), $^{**}$ ($p{<}.01$), and $^{***}$ ($p{<}.001$)) between memorization and compositionality scores across models and languages.}
\label{tab:mem-comp}
\end{table}


\begin{table}[ht]
\centering
\scriptsize
\setlength{\tabcolsep}{4pt}
\begin{tabular}{llll}
\toprule
\textbf{Model} & \textbf{Language} & \textbf{Point-biserial $r$} & \textbf{Spearman $\rho$} \\
\midrule
Aya-Expanse-32B      & EN & -0.067 *** & -0.067 *** \\
         & DE & -0.031 *   & -0.031 * \\
         & ZH & -0.025 *   & -0.026 * \\
         & KO &  0.017     &  0.017 \\
         & TR & -0.016     & -0.017 \\
\midrule
Qwen2.5-32B     & EN & -0.071 *** & -0.071 *** \\
         & DE & -0.061 *** & -0.061 *** \\
         & ZH & -0.074 *** & -0.074 *** \\
         & KO & -0.003     & -0.003 \\
         & TR & -0.020     & -0.020 \\
\midrule
DeepSeek-V3 & EN &  0.014     &  0.014 \\
         & DE & -0.041 **  & -0.042 ** \\
         & ZH & -0.023     & -0.023 \\
         & KO &  0.015     &  0.015 \\
         & TR & -0.059 *** & -0.060 *** \\
\midrule
GPT-4o   & EN & -0.052 *   & -0.052 * \\
         & DE & -0.045 **  & -0.045 ** \\
         & ZH & -0.070 *** & -0.071 *** \\
         & KO &  0.014     &  0.014 \\
         & TR & -0.043 *** & -0.043 *** \\
\bottomrule
\end{tabular}
\caption{Point-biserial $r$ and Spearman’s $\rho$ correlations and their statistical significance (marked with $^*$ ($p{<}.05$), $^{**}$ ($p{<}.01$), and $^{***}$ ($p{<}.001$)) between memorization and context across models and languages.}
\label{tab:mem-cont}
\end{table}

\begin{figure}[ht]
  \centering
  \includegraphics[width=0.9\linewidth]{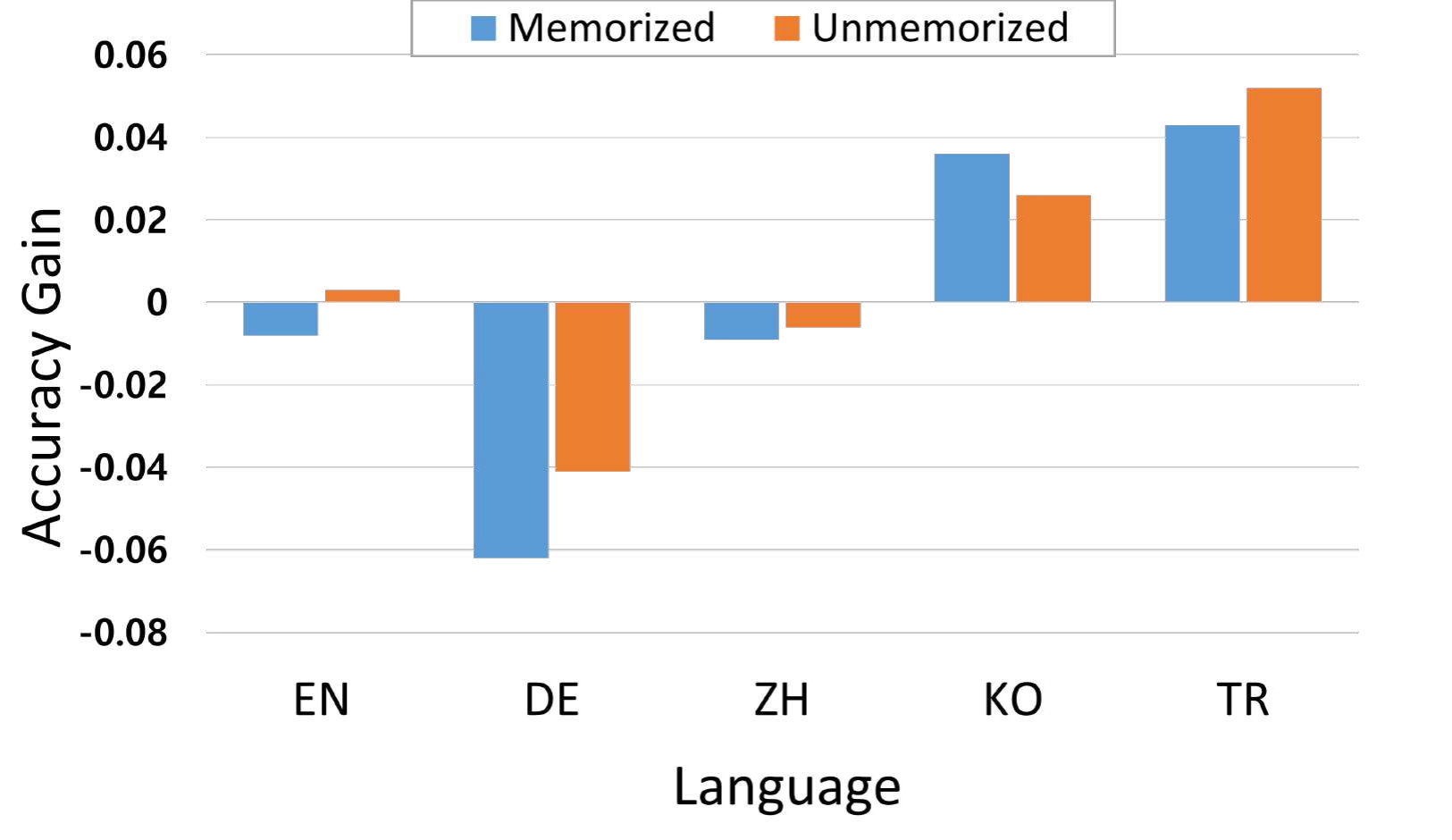}
  \caption{Accuracy gains from reasoning (i.e., the difference in performance with versus without reasoning) vary significantly across languages, obscuring the direct relationship between memorization and reasoning. These results are based on Qwen2.5-32B-Instruct.}
  \label{fig:mem-reason}
\end{figure}


\subsection{Factor Correlation Analysis}\label{appendix_factor}
This subsection describes the factor correlation analysis of \S\ref{main:factors} in detail. 
As mentioned, we consider five combinations with our factors paired in two: (1) memorization-compositionality, (2) memorization-context, (3) memorization-reasoning, (4) compositionality-context, and (5) compositionality-reasoning. 
We cover all four models for pairs without reasoning involved, and only Qwen for pairs that include reasoning. As for languages, five languages, excluding Arabic (due to the lack of context and unstable reasoning), are targeted, except for the memorization-compositionality pair, where we cover all languages. 

Out of the five factor pairs, we found correlations for two pairs: memorization and compositionality, along with memorization and context. 
In examining memorization and compositionality, we group idiom instances into five bins according to their compositionality scores and compare their memorization rate. As shown in Figure \ref{fig:mem-comp} and Table \ref{tab:mem-comp}, our findings indicate that LLMs are somewhat affected by compositionality, with statistically significant correlations across all model–language pairs, except for GPT-4o in Turkish.
For memorization with context, we compare the accuracy gains from additional context between memorized and unmemorized idiom groups across models and languages. As shown in Figure \ref{fig:mem-cont} and Table \ref{fig:mem-cont}, unmemorized groups often benefit more from added context, which can be interpreted as models compensating for limited internal knowledge by relying on contextual cues. However, the correlation between memorization and context is not as firm as the one between memorization and compositionality, with some model-language pairs exhibiting statistically insignificant correlations. 

For the other three pairs, we found no meaningful correlations.
More specifically, for pairs involving reasoning, language-specific inconsistencies persisted, obscuring the correlation between the two target factors, as illustrated in Figure \ref{fig:mem-reason}. 
Similarly, the interaction between compositionality-context varied substantially across languages, even though each factor individually showed a clear correlation with idiom understanding regardless of language.

\section{Appendix: Case Study Details}\label{appendix_case_study} 

\subsection{Works that Utilize LLMs for Idiom Processing}
\paragraph{Sentence Generation}
Works such as \citet{lee-2023} and \citet{liu-2024} use LLMs to generate context illustrating idiom usage, without explicitly providing its figurative meanings. This leaves the task entirely to the model’s internal understanding. \citet{liu-2024} highlights a critical limitation of this approach, noting that most examples generated by GPT-3.5 required manual revision due to incorrect idiom usage.

\paragraph{Machine Translation}
Works such as \citet{Li_2024} and \citet{donthi-2025} leverage IdiomKB to enhance the machine translation performance of relatively smaller LLMs. Specifically, they introduce KB-CoT, where the model is given a meaning of the target idiom extracted from IdiomKB to aid their translation process. However, all the included meanings in IdiomKB are entirely generated using GPT-3.5 without further verification efforts.

\paragraph{LLM-as-a-Judge}
\citet{Li_2024} and \citet{donthi-2025} use LLMs not only to enhance machine translation but also to evaluate the results. For instance, \citet{Li_2024} employs GPT-4 with prompts similar to Figure~\ref{fig:case_trans_eval} in the without-meaning setting. However, this approach is problematic, as it relies solely on GPT-4’s internal knowledge—an issue highlighted by Table~\ref{tab:mcq_results}, which shows suboptimal understanding for lower-resource languages.

\subsection{Number of Covered Instances}\label{case:stat}
The number of idiom instances covered in our case study is presented in Table~\ref{tab:case_instance}.

\begin{table}[h]
\centering
\small
\begin{tabular}{lcc}
\toprule
\textbf{Model} & \textbf{Language} & \textbf{\# Idioms} \\
\midrule
\texttt{Aya-Expanse-32B}         & KO & 982 \\
\texttt{Aya-Expanse-32B}         & TR & 1044 \\
\texttt{Qwen2.5-32B-Instruct}    & KO & 1068 \\
\texttt{Qwen2.5-32B-Instruct}    & TR & 1417 \\
\bottomrule
\end{tabular}
\caption{Number of covered idioms by model and language.}
\label{tab:case_instance}
\end{table}

\subsection{Machine Translation Results}
\begin{figure}[ht]
  \centering
  \includegraphics[width=0.85\linewidth]{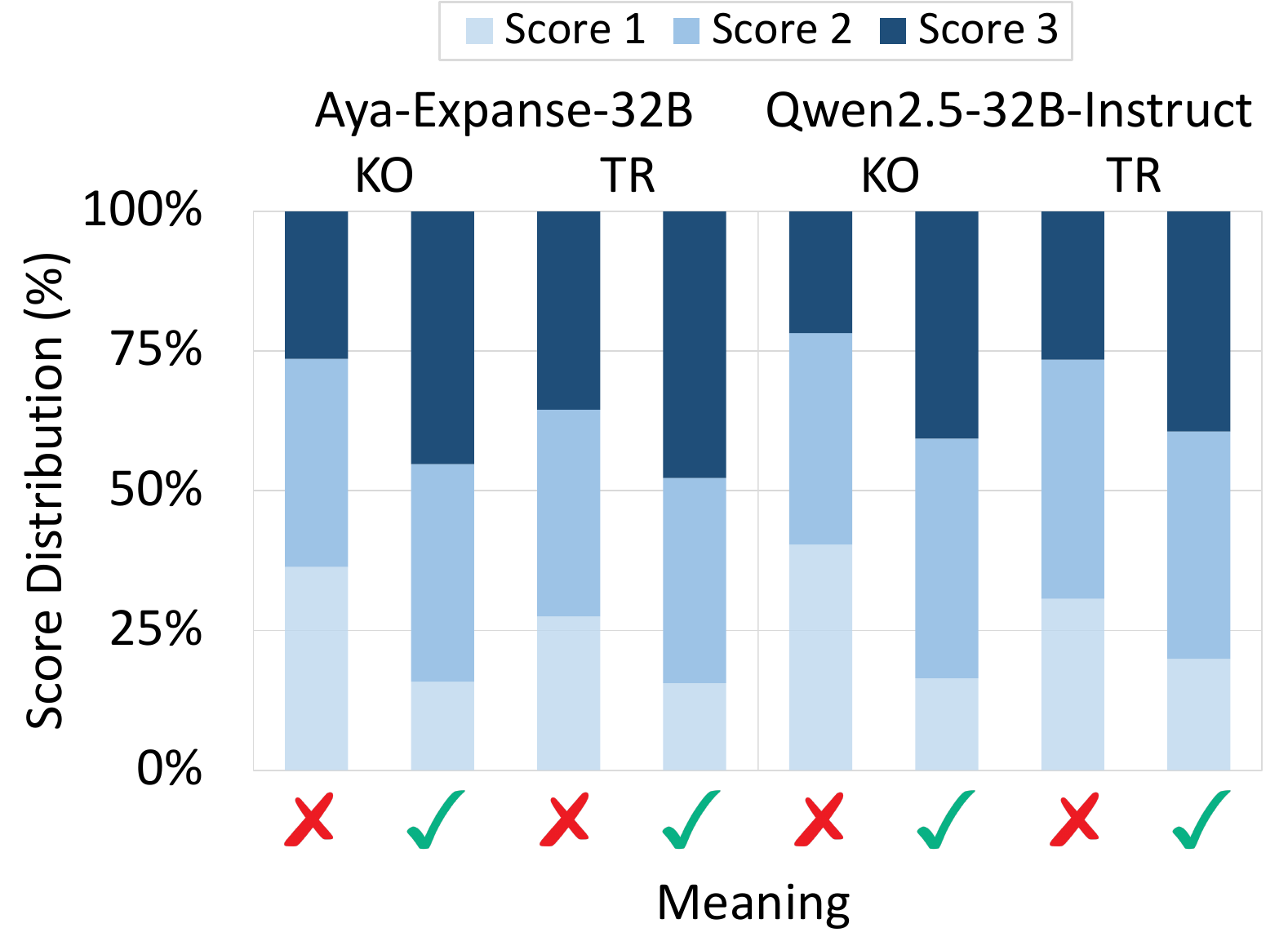}
  \caption{Scores assigned by GPT-4o to translated sentences generated by the models, evaluated under two conditions: when the idiom’s meaning was provided from our dataset (\cmark) and when it was generated by GPT-4o (\xmark).}
  \label{fig:case2}
\end{figure}

Results comparing the quality of machine-translated sentences with and without meanings are available in figure~\ref{fig:case2}. Although the gap between the two settings is smaller than in the sentence generation task, the results still clearly indicate that models perform better when idiom meanings are explicitly provided for machine translation as well.

\subsection{Prompt Details}\label{case:prompt}
This section introduces details about the prompts for our case study. Figure~\ref{fig:case_gen} includes both the Korean and Turkish prompts that we use to generate example sentences of idioms with and without meaning. Figure~\ref{fig:case_mean} is the prompt that we use to generate the meaning of the idioms using GPT-4o, which we provide in our following translation setting. Figure~\ref{fig:case_trans} is the prompt we use to translate Korean or Turkish sentences, including idioms, into English sentences. Figure~\ref{fig:case_gen_eval} and Figure~\ref{fig:case_trans_eval} are prompts that we use to evaluate the generated or translated sentences using GPT-4o.

\begin{figure}[ht]
  \setlength{\fboxsep}{7pt}
  \noindent
  \begin{tcolorbox}[
      colback=gray!10,       
      colframe=gray!75,      
      sharp corners=south,   
      rounded corners=northwest,
      boxrule=0.8pt,         
      width=\columnwidth,    
      fonttitle=\bfseries,
      coltitle=black,        
      title=Sentence Generation Prompt
  ]
    \fontsize{9pt}{11pt}\selectfont
    \setlength{\parskip}{5pt}

    \textcolor{orange}{\textbf{[Korean]}} \\
    주어진 관용표현과 그 의미를 바탕으로 한국어 예문 한 문장을 작성하세요:

    idiom: \texttt{\{idiom\}} \\
    \textcolor{cyan}{meaning: \texttt{\{meaning\}}} \\

    추가 설명 없이 예문만 출력하세요. 예문은 한국어로만 생성하세요. \\
    
    \textcolor{orange}{\textbf{[Turkish]}} \\
    Verilen deyim ve anlamı temel alarak bir örnek cümle oluşturun:

    idiom: \texttt{\{idiom\}} \\
    \textcolor{cyan}{meaning: \texttt{\{meaning\}}} \\

    Ek açıklama yapmadan sadece Türkçe cümleyi oluşturun. \\

     \textcolor{orange}{\textbf{[English (example)]}} \\
    Based on the given idiomatic expression and its meaning, write one example sentence in Korean/Turkish:

    idiom: \texttt{\{idiom\}} \\
    \textcolor{cyan}{meaning: \texttt{\{meaning\}}} \\

    Output only the example sentence without any additional explanation. The sentence should be generated in Korean/Turkish only. \\
    
  \end{tcolorbox}
  \caption{Prompts to generate example sentences of idioms for each language. We additionally provide the highlighted \textcolor{cyan}{meaning} part in the setting where the model has access to meanings. Note that the English prompt is not used in our sentence generation setting and is provided for illustrative purposes only.}
  \label{fig:case_gen}
\end{figure}

\begin{figure}[ht]
  \setlength{\fboxsep}{7pt}
  \noindent
  \begin{tcolorbox}[
      colback=gray!10,       
      colframe=gray!75,      
      sharp corners=south,   
      rounded corners=northwest,
      boxrule=0.8pt,         
      width=\columnwidth,    
      fonttitle=\bfseries,
      coltitle=black,        
      title=GPT-4o Meaning Generation Prompt
  ]
    \fontsize{9pt}{11pt}\selectfont
    \setlength{\parskip}{5pt}

    Given a \textcolor{orange}{Korean/Turkish} idiom, please write the idiom’s figurative \textcolor{orange}{Korean/Turkish} meaning. Please note: idioms always express figurative meaning that differs from the literal meaning of their constituent words.
    Return the meaning only—no extra words.
    
    Idiom: \texttt{\{idiom\}} \\
    Meaning:
    
  \end{tcolorbox}
  \caption{Prompt to generate idiom meanings with GPT-4o. The \textcolor{orange}{orange} part differs depending on source languages.}
  \label{fig:case_mean}
\end{figure}

\begin{figure}[ht]
  \setlength{\fboxsep}{7pt}
  \noindent
  \begin{tcolorbox}[
      colback=gray!10,       
      colframe=gray!75,      
      sharp corners=south,   
      rounded corners=northwest,
      boxrule=0.8pt,         
      width=\columnwidth,    
      fonttitle=\bfseries,
      coltitle=black,        
      title=Translation Prompt
  ]
    \fontsize{9pt}{11pt}\selectfont
    \setlength{\parskip}{5pt}

    "\texttt{\{idiom\}}" means \texttt{\{meaning\}}. \\

    Given the above knowledge, translate the following \textcolor{orange}{Korean/Turkish} sentence into English. \\
    Do NOT add any extra explanation. \\
  
    \textcolor{orange}{Korean/Turkish}: \texttt{\{sentence\}} \\
    English:"
    
  \end{tcolorbox}
  \caption{Prompt to translate sentences including idioms. The \textcolor{orange}{orange} part differs depending on source languages.}
  \label{fig:case_trans}
\end{figure}

\begin{figure}[ht]
  \setlength{\fboxsep}{7pt}
  \noindent
  \begin{tcolorbox}[
      colback=gray!10,       
      colframe=gray!75,      
      sharp corners=south,   
      rounded corners=northwest,
      boxrule=0.8pt,         
      width=\columnwidth,    
      fonttitle=\bfseries,
      coltitle=black,        
      title=Sentence Evaluation Prompt
  ]
    \fontsize{9pt}{11pt}\selectfont
    \setlength{\parskip}{5pt}

    You are an expert evaluator of idiom usage.
    Given an idiom\textcolor{cyan}{, its figurative meaning,} and an example sentence, rate how well the sentence reflects the idiom's figurative meaning on a scale of 1–3.
    Respond with only the number 1, 2, or 3.

    1: None of the given figurative meaning is conveyed
    2: Some of the given figurative meaning is conveyed  
    3: The given figurative meaning is fully and naturally conveyed

    Idiom: \texttt{\{idiom\}} \\
    \textcolor{cyan}{Figurative meaning: \texttt{\{figurative\_meaning\}}} \\
    Example sentence: \texttt{\{example\_sentence\}} \\

    Score (only number):
    
  \end{tcolorbox}
  \caption{Prompt to evaluate the generated sentences using GPT-4o. We additionally provide the highlighted \textcolor{cyan}{meaning} part in the setting where GPT-4o has access to meanings.}
  \label{fig:case_gen_eval}
\end{figure}

\begin{figure}[ht]
  \setlength{\fboxsep}{7pt}
  \noindent
  \begin{tcolorbox}[
      colback=gray!10,       
      colframe=gray!75,      
      sharp corners=south,   
      rounded corners=northwest,
      boxrule=0.8pt,         
      width=\columnwidth,    
      fonttitle=\bfseries,
      coltitle=black,        
      title=Translation Evaluation Prompt
  ]
    \fontsize{9pt}{11pt}\selectfont
    \setlength{\parskip}{5pt}

    You are an expert in idiom translations.
    /* Task prompt */
    Evaluate the idiom translation in the given English translation of a \textcolor{orange}{Korean/Turkish} sentence. Focus on the idiom’s figurative meaning.
    
    /* Evaluation Criteria */
    1 point: Ignores, mistranslates, or only translates the literal meaning of the idiom.
    2 points: Conveys basic figurative meaning but may lack refinement or have minor imperfections.
    3 points: Exceptional translation, accurately conveying figurative meaning, context, and cultural nuances.

    /* Test Data */
    Evaluate the following translation:
    \textcolor{orange}{Korean/Turkish} sentence: \texttt{\{src\}} \\
    Idiom in the \textcolor{orange}{Korean/Turkish} sentence: \texttt{\{idiom\}} \\
    \textcolor{cyan}{Idiom meaning: \texttt{\{meaning\}}} \\
    English translation: \texttt{\{tgt\}} \\
    Evaluation (score only):
    
  \end{tcolorbox}
  \caption{Prompt to evaluate the translated sentences using GPT-4o, following the format of evaluation from \citet{Li_2024}. We additionally provide the highlighted \textcolor{cyan}{meaning} part in the setting where GPT-4o has access to meanings. The \textcolor{orange}{orange} part differs depending on source languages.}
  \label{fig:case_trans_eval}
\end{figure}

\subsection{An Example of Generated Sentences}\label{case:ex}
Figure~\ref{fig:case_ex} shows an actual example of Korean sentences generated by Aya-Expanse-32B with and without access to the meaning of the target idiom.

\begin{figure}[ht]
  \setlength{\fboxsep}{7pt}
  \noindent
  \begin{tcolorbox}[
      colback=gray!10,       
      colframe=gray!75,      
      sharp corners=south,   
      rounded corners=northwest,
      boxrule=0.8pt,         
      width=\columnwidth,    
      fonttitle=\bfseries,
      coltitle=black,        
      title=Example of Generated Sentences
  ]
    \fontsize{9pt}{11pt}\selectfont
    \setlength{\parskip}{5pt}

    \textbf{ID:} 609-1

    \textbf{Idiom:} \textit{걸음발 타다} [Georeumbal tada] (literally “to ride the walking foot”)

    \textbf{Meaning:} 어린아이가 처음으로 걸음걸이를 익히다.  
    (“A young child learns to walk for the first time.”)

    \vspace{0.5em}
    \textbf{Generated sentence without providing meaning:}
    \begin{quote}
      \hangindent=1em
      그녀는 조용히 방에 들어와 걸음발을 타며 바닥에 쌓인 눈을 피해 조심스럽게 걸어갔다.\\
      (\textit{In English:} She quietly entered the room, \textit{[Georeumbal tada]}, and carefully walked to avoid the snow piled on the floor.)
    \end{quote}

    \vspace{0.5em}
    \textbf{With meaning provided:}
    \begin{quote}
      \hangindent=1em
      우리 막내가 걸음발을 타기 시작해서 온 집 안이 웃음바다가 되었다.\\
      (\textit{In English:} Our youngest started to \textit{[Georeumbal tada]}, and the whole house was filled with laughter.)
    \end{quote}
  \end{tcolorbox}
  \caption{Examples of Korean sentences generated by Aya-Expanse-32B with and without access to the meaning.}
  \label{fig:case_ex}
\end{figure}

\end{document}